\documentclass[journal]{IEEEtran}

\IEEEoverridecommandlockouts

\usepackage{amsmath,amsfonts}
\usepackage{algorithmic}
\usepackage{algorithm}
\usepackage{array}
\usepackage[caption=false,font=normalsize,labelfont=sf,textfont=sf]{subfig}
\usepackage{textcomp}
\usepackage{stfloats}
\usepackage{url}
\usepackage{verbatim}
\usepackage{graphicx}
\usepackage{cite}
\usepackage{amssymb}
\usepackage{mathnotation}
\usepackage{mathtools}
\usepackage[hidelinks]{hyperref}
\usepackage{gensymb}
\usepackage{tabularx}

\title{Geometric Data-Driven Multi-Jet Locomotion Inspired by Salps}

\author{Yanhao~Yang, Nina~L.~Hecht, Yousef~Salaman-Maclara, Nathan~Justus, Zachary~A.~Thomas, Farhan~Rozaidi, and Ross~L.~Hatton %
\thanks{This work was supported in part by ONR Award N00014-23-1-2171.} %
\thanks{All the authors are with the Collaborative Robotics and Intelligent Systems (CoRIS) Institute at Oregon State University, Corvallis, OR USA. {\tt\small \{yangyanh, Ross.Hatton\}@oregonstate.edu}}
}

\begin{document}

\maketitle

\begin{abstract}
Salps are marine animals consisting of chains of jellyfish-like units. Their efficient underwater locomotion by coordinating multi-jet propulsion has aroused great interest in robotics. This paper presents a geometric mechanics framework for salp-inspired robots. We study a new type of geometric mechanics models inspired by salps, in which control inputs are not restricted to the shape axes, analyze nonlinear controllability, and develop motion planning and feedback control methods. We introduce the ``LandSalp'' robot, which serves as a physical realization of the reduced-order, drag-dominated model of salp swimming, enabling controlled evaluation of locomotion strategies without many confounding factors of underwater experiments. We extend least-squares- and inverse-dynamics-based system identification to learn the Riemannian metric of the drag-dominated model from experimental data using Lie group differentiation. With about three minutes of data, we identify an accurate model of LandSalp. Simulation and hardware experiments demonstrate omnidirectional locomotion, shape regulation, and bending maneuvers, providing a principled pathway toward more capable salp-inspired robots.
\end{abstract}


\section{Introduction}
\label{sec:intro}

\begin{figure}[!t]
\centering
\includegraphics[width=\linewidth]{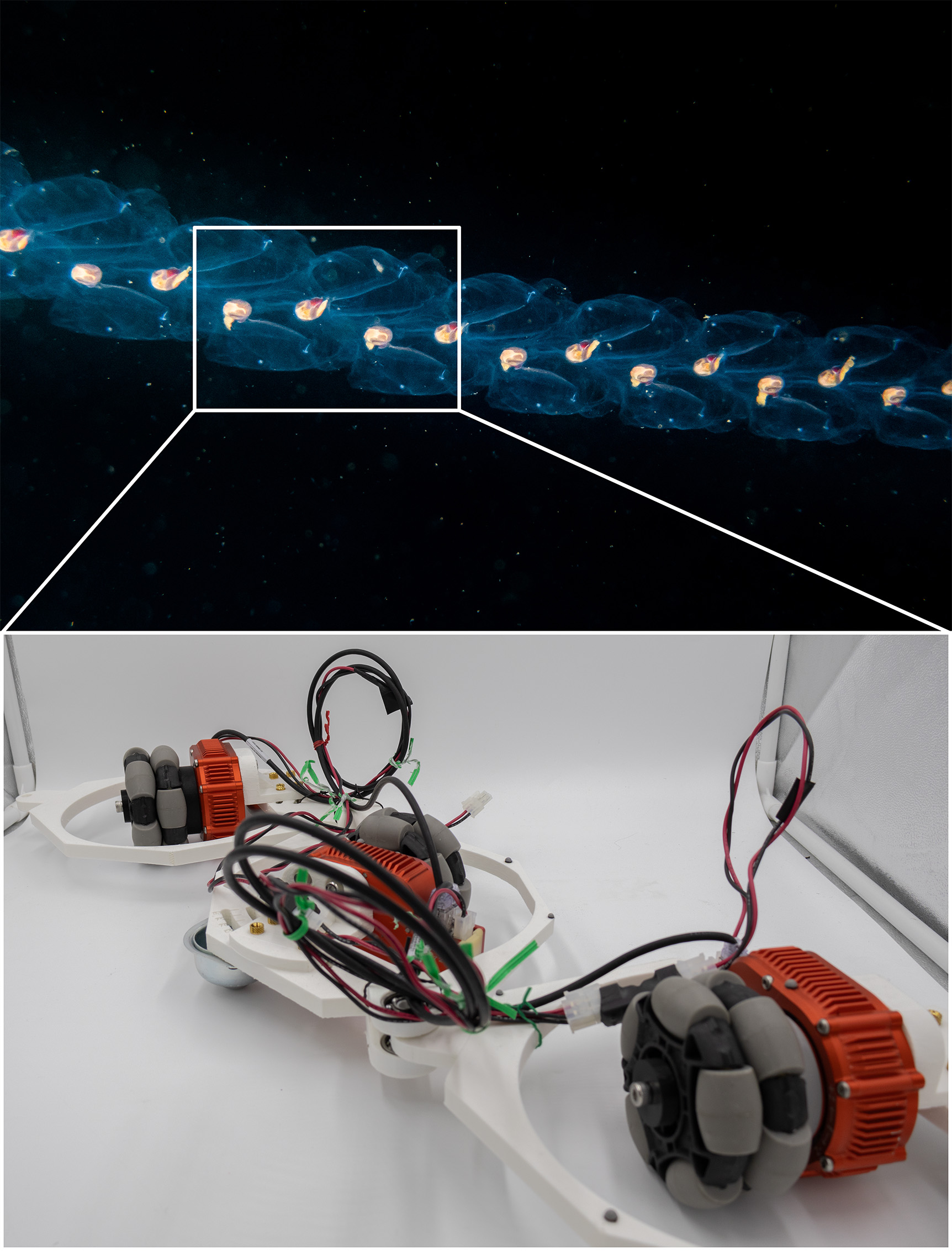}
\caption{The LandSalp robot (bottom) and biological salps (top). The LandSalp robot is a land-based physical realization of a reduced-order, drag-dominated model of biological salp, with a minimal but sufficient number of units for studying planning and control of salp-inspired robotic locomotion. The salp picture is reproduced from \cite{calliesandiego2024common}.}
\label{fig:intro}
\end{figure}

Salps are marine animals composed of chains of units\footnote{The units composing biological salps are called ``zooids'' (i.e., pseudoanimals or not-quite-animals) because they exhibit many properties of animals but are not independent organisms from the colony. To discuss the general properties of multi-jet locomotion without making claims about the biological systems that inspire them, we use the terminology ``chains'' and ``units'' throughout this paper.} resembling open-ended, jellyfish-like tubes. These creatures are capable of efficient underwater undulatory locomotion by coordinating multi-jet propulsion. The structure and locomotion patterns of salps are closely related, which has attracted widespread interest in both biological and ecological research \cite{bone1983jet, linden2004optimal, sutherland2010comparative, sutherland2017hydrodynamic, damian-serrano2024colonial}.

Salps and their jet-propulsion swimming have also inspired substantial work in robotics, spanning jet-propelled soft robot design \cite{renda2015structural, yang2021origamiinspired, jones2024underwater, dong2024salpot, yang2025effect, yang2025coupled} and multi-robot coordination \cite{aljarah2018asynchronous, ateya2019chaotic, rokbani2022beta, ding2022velocity, romeh2023multirobot}. However, much of the motion planning and control literature for undulatory locomotion assumes that control inputs are aligned with internal shape degrees of freedom, so that shape can be directly prescribed and used to generate propulsion \cite{ostrowski1996mechanics, mcisaac2003motion, morgansen2007geometric, hatton2013geometric, ramasamy2019geometry, hatton2022geometry}. In contrast, for salp-inspired multi-jet systems and related platforms, actuation is not generally aligned with shape coordinates: forces project into both shape and body motion, so both shape and position can be passive and cannot be cleanly decoupled under actuation \cite{tilbury1995trajectory, kimura2002development, takagi2023kinematicdual}. This coupling introduces new challenges for planning and control, since one must simultaneously achieve desired locomotion while maintaining shape stability---a concern that is particularly critical in hardware implementation.

Existing robotics studies of salps often address only a subset of the coupled design--model--control problem \cite{renda2015structural, yang2021origamiinspired, jones2024underwater, dong2024salpot, yang2025effect, yang2025coupled}. For example, many soft-actuator jet-propulsion designs successfully reproduce salp or jet-propulsion behaviors, but comparatively few works study how unit arrangement and jet orientation affect the resulting locomotion patterns and how to synthesize coordination and feedback control policies for chains of interacting jet units. Likewise, many multi-agent coordination methods assume each agent has low-level, near-instantaneous omnidirectional control authority \cite{aljarah2018asynchronous, ateya2019chaotic, rokbani2022beta, ding2022velocity, romeh2023multirobot}, an assumption that does not generally hold for jet-propelled chains where controllability and stability depend on the coupled shape--position dynamics.

This paper, based on geometric mechanics, presents a comprehensive study of salp-inspired robotic locomotion spanning mechanism design, dynamic modeling, system identification, motion planning, and feedback control. We introduce a land-based robotic platform, ``LandSalp'', as an extension of previous work \cite{hecht2024design}. LandSalp serves as a physical realization of a reduced-order, drag-dominated model of salp swimming, enabling controlled evaluation of planning and control strategies while reducing confounding factors common in underwater experiments. The reduced-order model captures the core dynamic features of interest: chain-like locomotion with thrust directions not aligned with joint axes in a viscous, drag-dominated regime. These results contribute toward a principled foundation for salp-inspired robot design and control, and they provide a pathway for extending the framework to more capable underwater robotic systems.

Specifically, the contributions of this work are as follows:
\begin{enumerate}
\item We investigate a new geometric mechanics model in which control is not restricted to the shape coordinates, resulting in passive position and shape dynamics in a viscous, drag-dominated regime;
\item We analyze nonlinear controllability and propose geometric motion planning and feedback control methods for this highly underactuated locomoting system;
\item We extend least-squares- and inverse-dynamics-based identification techniques to learn the Riemannian metric of a drag-dominated dynamic model from experimental motion data using Lie group differentiation;
\item We integrate geometric mechanics with biological observations of low-Reynolds-number swimmers to study salp-inspired locomotion, and we present the LandSalp robot as a land-based physical platform for testing these methods with minimal confounding factors.
\end{enumerate}
Based on the proposed data-driven geometric modeling approach, we successfully developed an accurate dynamic model of the LandSalp robot with only about 3 minutes of experimental data. Utilizing this model, we applied the proposed geometric motion planning and feedback control algorithms to demonstrate the omnidirectional locomotion of LandSalp in both simulated environments and physical hardware.

\section{Geometric Modeling for Underwater Locomotion of Salps}
\label{sec:model_and_design}

\begin{table}[!t]
    \caption{Key Symbols Used Throughout This Paper}
    \centering
    \begin{tabularx}{\linewidth}{l X}
    \hline \hline
        \textbf{Symbol} & \textbf{Meaning} \\
    \hline
        $\jointangle, \dot{\jointangle}$ & Individual joint positions and velocities \\
        $\beta_i$ & Orientation of the $i$-th jet thrust relative to the corresponding link axis \\
        $\base, \basedot$ & System shape and shape velocity \\
        $\fiber, \fibercirc$ & System position and body velocity \\
        $\bundle, \bundledot$ & System configuration and its (group) derivatives \\
        $\ctrl$ & Controlled jet force or controlled jet velocity \\
        $\coforce, \lagrangeforce$ & Forces and moments expressed in the system's body frame and joint torques \\
        $\tildecoforce$ & Force measurement from the series-elastic actuator \\
        $\force_{\bundle}$ & Net external forces and moments pulled back to the system configuration coordinates \\
        $\localmetrics^{\mathrm{unit}}, \localmetrics^{\mathrm{joint}}, \Localmetrics$ & Salp unit drag coefficient (expressed in local body coordinates), joint viscous friction coefficient, and the aggregate drag coefficient matrix \\
        $\jac_{\fstidx}$ & Jacobian matrix mapping system configuration velocities to the body velocity of the $\fstidx$-th salp unit \\ 
        $\jac_{\bundle}, \jac_{\ctrl}$ & Jacobians mapping configuration velocity and controlled jet velocity, respectively, to the body velocity of the salp units and joints, where viscous drag occurs and drag coefficients are defined \\
        $\jac_{\force_{\bundle}}, \jac_{\coforce}$ & Jacobian matrices mapping system drag coefficients to net external forces and moments in system configuration coordinates and series-elastic actuator directions \\ 
        $\cvf$ & Control vector field \\
        $\selection$ & Selection matrix extracting the force component aligned with the series-elastic actuator \\
        $\altfiber$ & Rigid body transformation within the system \\
        $\lambda$ & Regularization strength \\
        $\bar{\ctrl}, A^{\mathrm{sin}}, A^{\mathrm{cos}}$ & Fourier series constant and oscillation coefficients \\
        $\Ctrl$ & Control parameters \\
        $T, \omega$ & Gait period and frequency \\
        $R$ & Gait cost metric \\
        $K$ & Feedback gains \\        
        $(\cdot)^{\mathrm{aug}}$ & Augmented terms with Lie brackets accounting for gait cycle effects \\
        $(\cdot)^{\mathrm{desired}}$ & Desired motion for control \\
    \hline \hline
    \end{tabularx}
    \label{tab:symbol}
\end{table}

In this section, we begin with a brief historical perspective on geometric mechanics in locomotion, and then specialize this framework to salp-inspired multi-jet locomotion. We model a salp chain as a set of rigid links connected by passive joints and driven by distributed jets, and we assume a low-Reynolds-number regime in which viscous drag dominates. Under this assumption, quasi-static force balance yields first-order dynamics that can be written as a control vector field mapping jet forces to configuration velocities.

\subsection{History of Geometric Mechanics for Locomotion}
Locomotion is inherently relative: it concerns changes in a body's pose in the world, which naturally live on Lie groups such as $\mathrm{SE}(2)$ or $\mathrm{SE}(3)$ rather than in Euclidean spaces. Geometric mechanics provides a systematic framework, grounded in Lie groups and Lie algebras, for describing this relative motion, deriving reduced-order models, and analyzing controllability and motion planning. For clarity, generality, and connection to existing literature, we adopt this framework explicitly rather than re-deriving equivalent structures in an ad hoc manner. Geometric mechanics has enabled concise modeling and analysis across diverse locomotion modalities through local metrics \cite{yang2024geometric}, including low- and high-Reynolds-number swimming \cite{hatton2013geometric}, terrestrial wheel-driven motion \cite{hatton2011geometric}, and self-reorientation in space \cite{walsh1995reorienting}, and it has supported advances in robot design \cite{rozaidi2023hissbot}, modeling \cite{hatton2013geometric, sparks2022amoebainspired}, system identification \cite{bittner2018geometrically, sparks2022amoebainspired, even2024variational}, and motion planning \cite{ramasamy2019geometry, hatton2022geometry}. Building on these insights, we use geometric mechanics to obtain a compact, reduced-order model capturing the key physics of salp-inspired multi-jet locomotion, to learn the Riemannian metric characterizing drag-dominated dynamics from minimal experimental data, and to extend geometric motion planning and feedback control methods to settings in which actuation is not confined to the shape coordinates.

\begin{figure}[!t]
\centering
\includegraphics[width=\linewidth]{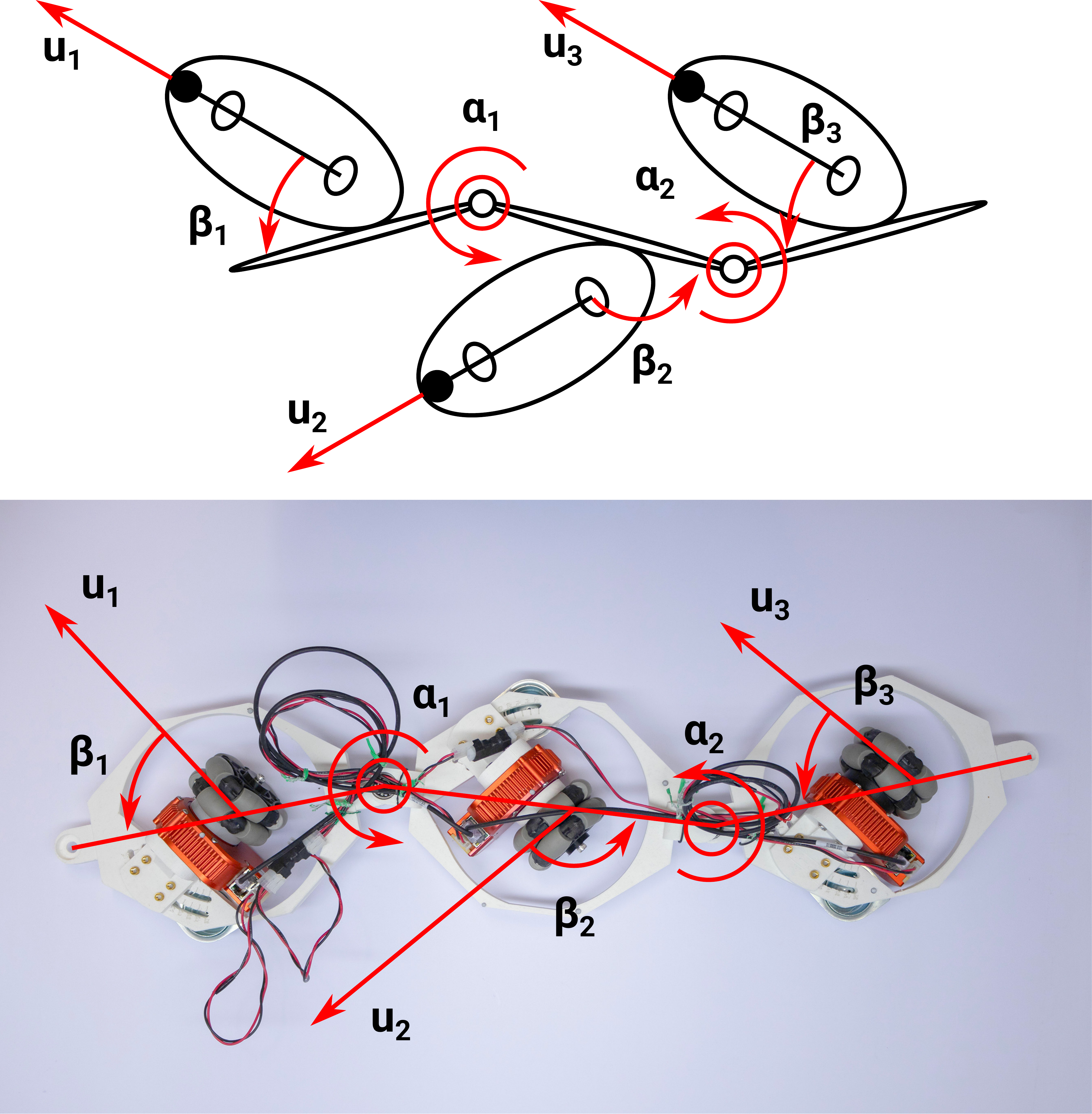}
\caption{Schematic comparison of the LandSalp robot (bottom) and the corresponding biological salp with linear-like architectures (top). There are two joints ($\alpha_1$ and $\alpha_2$) connecting the links from left to right. The orientations of the three motor-wheel assemblies used to generate jet thrust and (part of the) viscous drag are $\beta_1$, $\beta_2$, and $\beta_3$, respectively, from left to right, measured relative to the corresponding link axis.}
\label{fig:configuration}
\end{figure}

\subsection{Kinematics}

A planar salp chain with $\fstqty$ units can be modeled as a chain of $\fstqty$ links with $\fstqty-1$ passive joints and $\fstqty$ point sources of thrust (corresponding to the water-jet siphons), with jet directions alternating toward the left-rear and right-rear of the chain (Fig.~\ref{fig:configuration}).\footnote{In this paper, we focus on planar motion of salps with linear architectures for clarity. However, neither the modeling procedure nor the control methods rely on properties unique to planar motion or linear chains. Other architectures can be handled by considering a more general kinematic tree, and three-dimensional motion can be modeled by replacing $\mathrm{SE}(2)$ with $\mathrm{SE}(3)$. While $\mathrm{SE}(3)$ introduces higher-dimensional and more strongly non-commutative rotations, our methods already account for the non-commutativity between rotation and translation on $\mathrm{SE}(2)$, and the underlying geometric constructions (e.g., body velocities, connections, and Riemannian metrics) generalize directly to $\mathrm{SE}(3)$.} In this model, the control input is the jet force $\ctrl$, and the passive configuration velocity can be decomposed\footnote{Formally, this decoupling is embodied in a differential geometric structure called a \emph{fiber bundle}, where the \emph{shape} and \emph{position} of the system are respectively referred to as elements of the \emph{base} and \emph{fiber} spaces \cite{bloch2015nonholonomic}.} into the body velocity\footnote{The ``open circle'' notation we use here is similar to the ``dot'' notation, but denotes velocities expressed with local forward and lateral components rather than coordinate-aligned components. Please refer to Appendix~\ref{appendix:group} for more details.}\textsuperscript{,}\footnote{To improve accuracy of the averaging approximation mentioned in Section~\ref{sec:motion_planning} and better represent the robot as a whole, we use a body frame defined by the center of mass and the mean orientation of each link \cite{hatton2009approximating}. For a more detailed discussion on the choice of coordinates that define the system, please refer to Appendix~\ref{appendix:coordinate}.} of the system~$\fibercirc$ and shape (joint) velocities~$\basedot$
\beqbracket
    \ctrl = \begin{bmatrix}
    \ctrl_{1} \\ \vdots \\ \ctrl_{\fstqty}
    \end{bmatrix} \quad \basedot = \begin{bmatrix}
    \dot{\jointangle}_{1} \\ \vdots \\ \dot{\jointangle}_{\fstqty-1}
    \end{bmatrix} \quad
    \bundledot = \begin{bmatrix}
        \fibercirc \\ \basedot
    \end{bmatrix}. \label{eq:def}
\eeqbracket
The body velocity of each salp unit, $\fibercirc_{\fstidx}$, can be computed from the configuration velocity using differential kinematics, given the transformation from the system body frame to the frame of the $\fstidx$-th salp unit $\altfiber_{\fstidx}(\base)$ and the transformation from the $\sndidx$-th joint to the $\fstidx$-th salp unit $\altfiber_{\frac{\fstidx}{\sndidx}}(\base)$
\begin{eqalign} \label{eq:kinematics}
    \fibercirc_{\fstidx} &= \inv{\Adj_{\altfiber_{\fstidx}}}\fibercirc + \sum^{\fstidx-1}_{\sndidx = 1}\inv{\Adj_{\altfiber_{\frac{\fstidx}{\sndidx}}}}\begin{bmatrix}0 \\ 0 \\ \dot{\jointangle}_{\sndidx}\end{bmatrix} \\ 
    &= \jac_{\fstidx}\bundledot,
\end{eqalign}
where $\jac_{\fstidx}$ is the Jacobian that maps the configuration velocity to the body velocity of the $\fstidx$-th salp unit and $\Adj$ is the adjoint operator.\footnote{The adjoint operator $\Adj_{\bullet}$ combines the cross-product operation---which converts linear and angular velocities or forces/momenta between different frames---with the rotation operation that expresses velocities or forces/momenta in body-aligned coordinates.}

\subsection{Dynamics}

Unlike many geometric mechanics models for locomotion that assume control aligned with the shape coordinates \cite{ostrowski1996mechanics, mcisaac2003motion, morgansen2007geometric, hatton2013geometric, ramasamy2019geometry, hatton2022geometry}, salp-inspired multi-jet locomotion applies thrust that is generally off-axis relative to the shape coordinates. Jet forces therefore project into both shape and body-motion directions, yielding coupled, passive position and shape dynamics. To model this regime, we build on biological observations of salps swimming at low Reynolds numbers \cite{sutherland2017hydrodynamic} and derive a reduced-order, viscous drag-dominated model.

Biological analysis of salps indicates that they swim at relatively low Reynolds numbers \cite{sutherland2017hydrodynamic}. This condition allows us to reasonably set up the reduced-order model for the salp-inspired robotic locomotion as being dominated by viscous drag, with any inertial effects quickly damped out when the system motion frequency is far below the frequency associated with the system time constant (i.e., inertia divided by drag) \cite{hatton2013geometric}.\footnote{To extend the proposed methods beyond this assumption, please refer to \cite{hatton2022geometry} for an inertia-dominated kinematic model, to \cite{justus2024optimal} for a dynamic model that considers both the inertial and drag effects on the body as well as the spring stiffness at the joints, and \cite{yang2025geometricb} for motion planning for kinodynamic systems. 

Biological salps are also influenced by effects such as added mass, unsteady vortex shedding, and quadratic drag; these can be treated as disturbances to the feedback controller used here or incorporated explicitly in future extensions.} This model has two consequences for our system analysis:
\begin{enumerate}
\item The drag on each salp unit is a linear function of its body velocity, which can be pulled back to be a linear function of the body and shape velocity of the entire system; and
\item An isolated salp chain is in quasi-static equilibrium: the net drag forces and moments from its interaction with the surrounding fluid and joint frictions are equal and opposite to the net forces and moments induced by the jet.
\end{enumerate}

The first consequence of the low Reynolds number assumption (linear drag) allows us to derive the net external force and moment on each salp unit as the sum of jet thrust and viscous drag:
\beq
    \coforce_{\fstidx} = \ctrl_{\fstidx} - \localmetrics^{\mathrm{unit}}_{\fstidx}\fibercirc_{\fstidx},
\eeq
where the subscript $\fstidx$ denotes the $\fstidx$-th salp unit, and all terms are expressed in that unit’s body frame: $\coforce_{\fstidx}$ represents the net external force and moment on the unit, $\ctrl_{\fstidx}$ denotes the jet force, $\localmetrics^{\mathrm{unit}}_{\fstidx}$ is the drag coefficient matrix, and $\fibercirc_{\fstidx}$ represents the body velocity of the unit, which can be calculated using~\eqref{eq:kinematics}. 
This formulation makes it possible to simulate the reduced-order model of underwater locomotion of biological salps using the LandSalp robot with precise series-elastic force control. 

Similarly, we also model the viscous friction on each joint as being proportional to its rotational velocity:
\beq
    \jointtorque_{\sndidx} &= -\localmetrics^{\mathrm{joint}}_{\sndidx}\dot{\jointangle}_{\sndidx},
\eeq
where the subscript $\sndidx$ denotes the $\sndidx$-th joint, and all terms are expressed in the generalized coordinates of the $\sndidx$-th joint: $\jointtorque_{\sndidx}$ represents the viscous friction on the joint, $\localmetrics^{\mathrm{joint}}_{\sndidx}$ is the drag coefficient, and $\basedot_{\sndidx}$ represents the joint velocity.

\begin{figure}[!t]
\centering
\includegraphics[width=\linewidth]{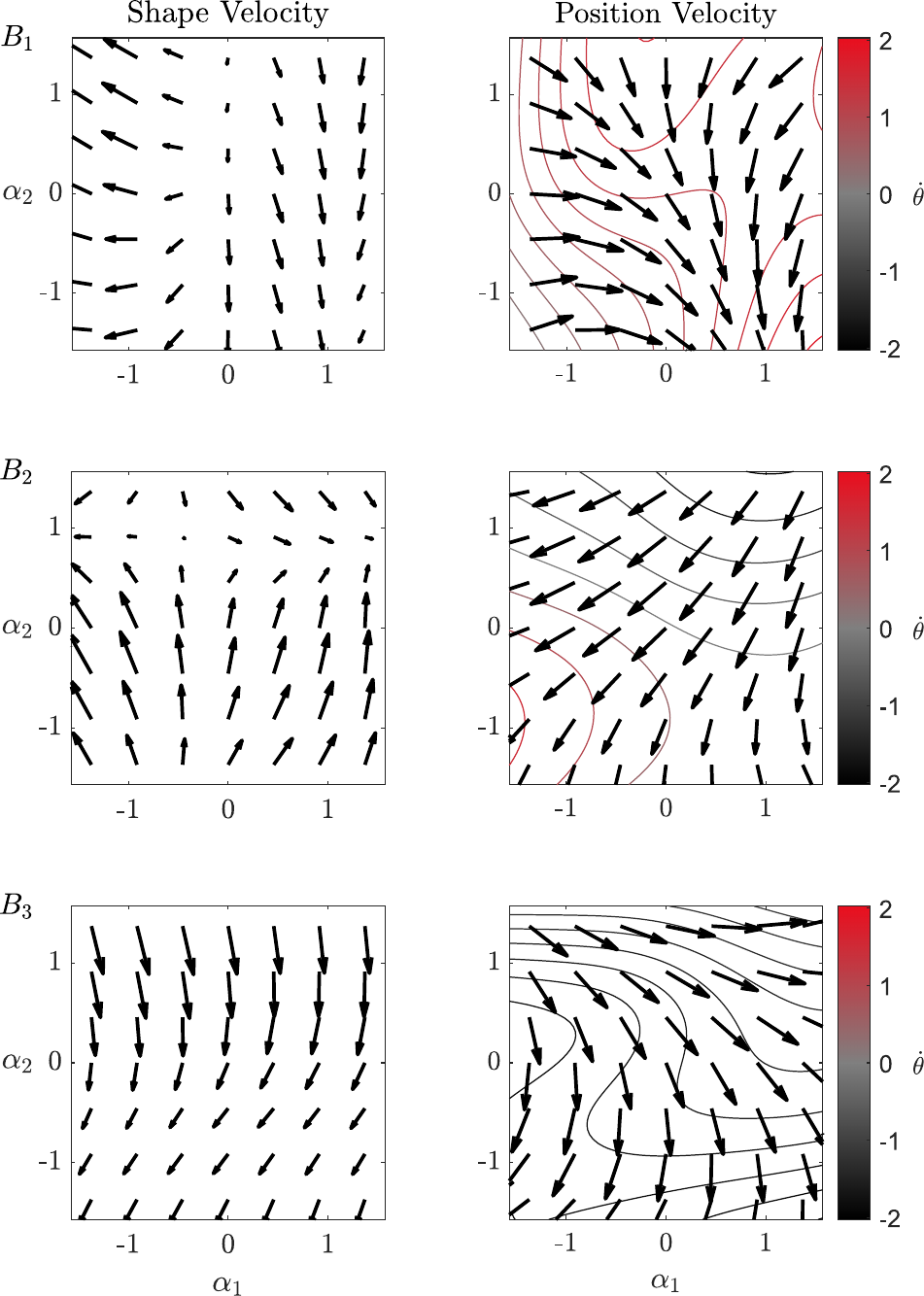}
\caption{Control vector fields $\cvf$ for the LandSalp data-driven model. Each row corresponds to each control input. The left column shows the shape velocity resulting from each control input at different shapes, and the right column shows the position velocity resulting from each control input at different shapes, with the linear velocity in the $\dot{x}$ and $\dot{y}$ directions overlaid with the shape axes, and the contours showing the angular velocity $\dot{\theta}$.}
\label{fig:cvf}
\end{figure}

The second consequence of the low Reynolds number assumption (quasi-static equilibrium) provides the nonholonomic constraints for determining the system motion based on the controlled jet force. These nonholonomic constraints can be expressed as the condition that the net external force and moment applied to the system configuration coordinates sum to zero:
\begin{eqalign}
    \force_{\bundle} &=  \transpose{\begin{bmatrix}\transpose{\coforce} & \transpose{\force}_{\base}\end{bmatrix}} \\
    &= \sum^{\fstqty}_{\fstidx}\transpose{\jac}_{\fstidx}\coforce_{\fstidx} + \begin{bmatrix}
        \boldsymbol{0} \\ \boldsymbol{\jointtorque} 
    \end{bmatrix} = \boldsymbol{0} \label{eq:min_energy}
\end{eqalign}
where $\jac_{\fstidx}$ is the Jacobian that maps the configuration velocity to the body velocity of the $\fstidx$-th salp unit, and its transpose maps the forces acting on the salp unit back to the configuration coordinates. The term $\force_{\bundle}$ includes the net external force and moment applied to the system's body frame, as well as the net external moment acting on the joints. Specifically, this term is computed by pushing forward the total forces and moment on each salp unit, $\coforce_{\fstidx}$, using the corresponding Jacobians and incorporating the effects of viscous friction at each joint, $\boldsymbol{\jointtorque}$.

The net external forces and moments acting on the salp system include the viscous drag on each salp unit and joint, as well as the active water jet force on each salp unit. These forces and moments can all be mapped to the configuration coordinates using the corresponding Jacobians, resulting in the force balance equation given in~\eqref{eq:min_energy}:
\begin{eqalign} 
    \force_{\bundle} &= \sum^{\fstqty}_{\fstidx}\transpose{\jac}_{\fstidx}(\ctrl_{\fstidx} - \localmetrics^{\mathrm{unit}}_{\fstidx}\fibercirc_{\fstidx}) - \begin{bmatrix}
        \boldsymbol{0} \\ \localmetrics^{\mathrm{joint}}_{1}\dot{\jointangle}_{1} \\ \vdots \\ \localmetrics^{\mathrm{joint}}_{\fstqty-1}\dot{\jointangle}_{\fstqty-1}
    \end{bmatrix} \\
    &= \transpose{\jac_{\bundle}}(\jac_{\ctrl}\ctrl + \Localmetrics\jac_{\bundle}\bundledot), \label{eq:pullback}
\end{eqalign}
where $\jac_{\bundle}$ is the Jacobian that maps the configuration velocity to the body velocity of the salp units and the joints where viscous drag occurs and the drag coefficients are defined; matrix $\jac_{\ctrl}$ selects and rotates the commanded jet forces into each unit's body frame; and $\Localmetrics$ denotes the aggregated local metrics, represented as a block diagonal matrix that contains the drag coefficients for each salp unit and joint. 
By setting the total forces and moments on the configuration coordinates to zero, one can solve the system's equation of motion \cite{yang2024geometric}
\beq \label{eq:dynamics}
    \bundledot = \underbrace{-\inv{(\transpose{\jac}_{\bundle}\Localmetrics\jac_{\bundle})}\transpose{\jac}_{\bundle}\jac_{\ctrl}}_{\cvf}\ctrl
\eeq
where $\cvf$ is the control vector field that maps control to the body and shape velocity of the system. Fig.~\ref{fig:cvf} illustrates the control vector fields of the LandSalp robot we designed to emulate the underwater locomotion of salps. The following section discusses this design in detail.

\section{LandSalp Design for Simulating Underwater Locomotion of Salps}

\begin{figure}[!t]
\centering
\includegraphics[width=\linewidth]{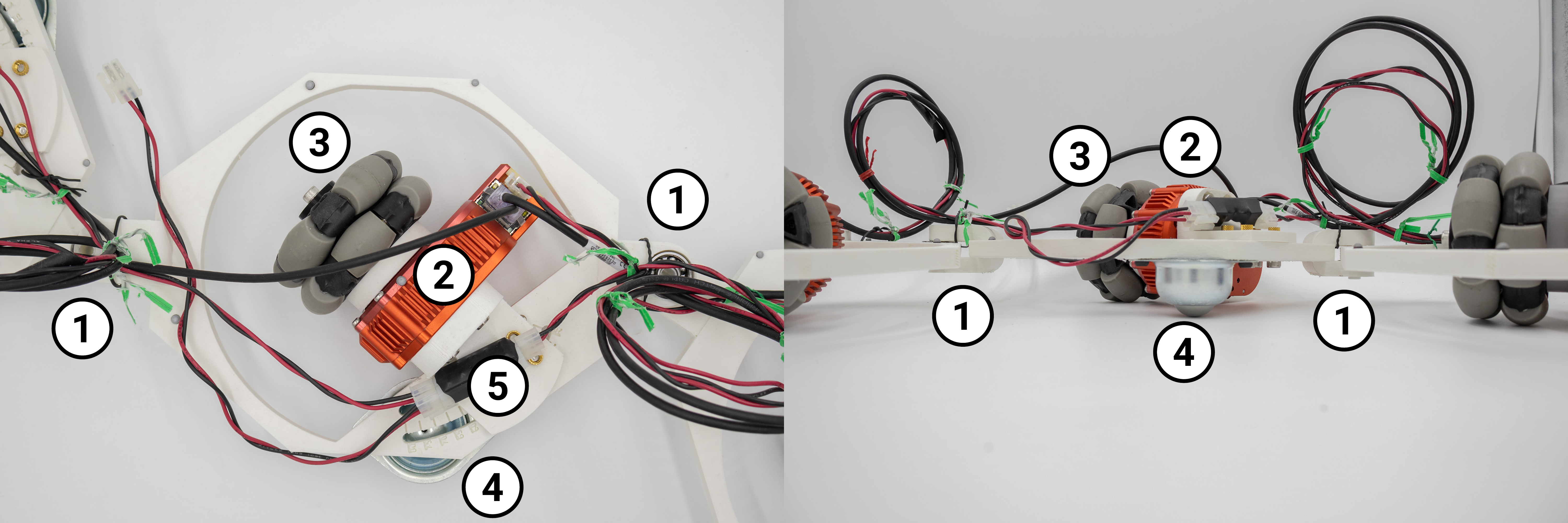}
\caption{The LandSalp is constructed as a planar chain of links, each link consisting of the following components: 1) passive joints, 2) HEBI X5 series-elastic actuators, whose output torque is a combination of a commanded driving torque and a drag term that simulates fluid friction acting on the link as it moves, 3) omniwheels, 4) ball casters for balancing, 5) rotation mounting slots for adjusting wheel orientation to correspond to different jet directions.}
\label{fig:design}
\end{figure}

Considering that the underwater locomotion of salps is at relatively low Reynolds numbers \cite{sutherland2017hydrodynamic}, for which viscous drag can dominate the dynamics, the total force and moment exerted on each salp unit are approximately proportional to the difference between the controlled jet force and the salp unit's body velocity, as demonstrated in~\eqref{eq:forceconvert}. This relationship allows us to design the LandSalp robot, which leverages the precise force control of series-elastic actuators, to physically realize the reduced-order, drag-dominated model and to reproduce the key dynamic characteristics of salp-inspired locomotion---namely, chain-like motion with thrust not aligned with joint axes and passive joint angles in a viscous, drag-dominated regime, on a flat table surface without the testing overhead of aquatic operations. LandSalp allows us to establish our understanding of the modeling, planning, and feedback control strategies of a chain-like, salp-inspired robot subjected to viscous drag and equipped with distributed thrust, with a minimum of confounding factors, while also providing a clear development pathway for extending these strategies to more complicated and capable swimming systems.

The LandSalp is constructed as a planar chain of links connected by passive joints, as illustrated on the bottom side of Fig.~\ref{fig:intro} and on the bottom side of Fig.~\ref{fig:configuration}. As illustrated in Fig.~\ref{fig:design}, each link is equipped with a HEBI series-elastic actuator driving an omniwheel, and has a ball-caster acting as a balance support, ensuring that the links remain in the horizontal plane. Each of these links forms a unit in our LandSalp robot (corresponding to the zooids in a biological salp), with the motor-omniwheel assemblies simulating the water jet propulsion and viscous drag, and the omniwheel--caster combination reduces Coulomb friction to support the drag-dominated modeling assumption.

\begin{table}[!t]
\begin{center}
\caption{Specifications of LandSalp}
\label{tab:pars}
\begin{tabular}{l | l}
\hline
\hline
Parameter & Value \\
\hline
Link Length (m) & $0.27$ \\
Omniwheel Diameter (m) & $0.083$ \\
Omniwheel Orientation (deg) & $-57$, $-130$, $-57$ \\
Actuator Maximum Speed (RPM) & $32$, $14$, $32$ \\
Actuator Continuous Torque (Nm) & $4$, $8$, $4$ \\
\hline 
\hline 
\end{tabular}
\end{center}
\end{table}

The motor-omniwheel assemblies are individually controlled such that their output torque is a combination of a commanded driving torque and a drag term that simulates fluid drag acting on the link as it moves. In addition, viscous friction arises from the omniwheel rollers, the wheel steering, the ball caster motion, and the joint rotations. Notably, the robot’s actuation and drag closely mimic the force distribution observed in salps, where the net forces comprise both the water jet force and viscous drag.

Furthermore, the constrained joint angles require the motors to operate the wheels in a cyclical and reciprocating fashion - referred to as a gait in Section~\ref{sec:motion_planning} - to propel the robot, which essentially simulates the pulsating intake and ejection of water observed in salps.\footnote{Note that in this paper, we assume that the LandSalp robot is capable of bidirectional wheel actuation. As discussed in~\cite{sutherland2010comparative}, salps have incurrent and excurrent siphons at opposite ends of their bodies, enabling reverse swimming---albeit with different efficiencies. Future work will explore unidirectional jet propulsion, which poses additional control challenges due to its inherent limitations in instantaneous controllability. Such limitations may be overcome by implementing a cyclic gait whose force offsets yield a net thrust bias, as discussed in Section~\ref{sec:motion_planning}.} These assemblies can be placed at different orientations relative to the link by being mounted in different positions of the rotation slot, corresponding to different orientations of the jets on a salp relative to its body and are parameterized by the angles $\beta_1$, $\beta_2$, and $\beta_3$. For linear-like salp chains, jet directions typically alternate toward the left-rear and right-rear of the body, as shown in the top of Fig.~\ref{fig:configuration}. Table~\ref{tab:pars} shows the main specifications of the LandSalp robot. It is worth noting that we used a different motor module on the central link, which has approximately half the maximum speed and twice the torque limit compared to the motors on the side links. This difference in motors further breaks the symmetry of the system and better reflects practical scenarios such as partial module failure or cooperation among heterogeneous modules. Importantly, both the thrust directions (relative to each link) and the actuator limits are explicitly parameterized in our model and controllers, so the resulting design and control pipeline applies beyond this particular hardware configuration.

Once the kinematics are updated according to the dimensions of LandSalp, the geometric modeling framework derived from Section~\ref{sec:model_and_design} applies directly. It should be noted that in the hardware implementation, the low-level command to the series-elastic actuators is a jet \emph{velocity} rather than a jet \emph{force}. This velocity command can be related to jet force by normalizing by the drag coefficient, yielding, yielding the total force and moment on each salp unit
\beq
    \coforce_{\fstidx} = \localmetrics^{\mathrm{unit}}_{\fstidx}\left(\ctrl_{\fstidx} - \fibercirc_{\fstidx}\right), \label{eq:forceconvert}
\eeq
where $\ctrl_{\fstidx}$ denotes the commanded jet velocity. With the updated Jacobian $\jac_{\ctrl}$, the total forces and moments acting on the LandSalp becomes
\beq
    \force_{\bundle} = \transpose{\jac_{\bundle}}\Localmetrics\begin{bmatrix}
        \jac_{\bundle} & \jac_{\ctrl}
    \end{bmatrix}\begin{bmatrix}
        \bundledot \\
        \ctrl
    \end{bmatrix},
\eeq
and the first-order dynamics are
\beq
    \bundledot = \underbrace{-\inv{(\transpose{\jac}_{\bundle}\Localmetrics\jac_{\bundle})}\transpose{\jac}_{\bundle}\Localmetrics\jac_{\ctrl}}_{\cvf}\ctrl.
\eeq
For simplicity, in the remainder of the paper we use $\ctrl$ to denote jet velocity to align with the LandSalp implementation. The same framework generalizes directly to platforms where thrusters or propellers are controlled via jet force.

\section{Data-Driven Geometric Model Identification}

Although Section~\ref{sec:model_and_design} establishes a reduced-order geometric model for salp-inspired locomotion in a viscous, drag-dominated regime, the drag coefficients that parameterize this model remain unknown. Rather than estimating these coefficients from CAD-based dimensions or tuning them by hand, we learn the metric directly from experimental data using a structure-preserving inverse-dynamics formulation on the Lie group. This approach is both more accurate and more scalable in practice, and it can adapt to changes in environment (e.g., surface conditions or fluid properties) through re-identification.

Specifically, we extend least-squares and inverse-dynamics-based system identification methods to identify the Riemannian metric of a drag-dominated locomotion model from experimental trajectories. We perform this identification using Lie-group differentiation, thereby respecting the geometry of the configuration space and exploiting the structure provided by Lie groups and Lie algebras. The main assumption underlying this framework is that, consistent with the modeling in Section~\ref{sec:model_and_design}, the system dynamics are dominated by viscous drag encoded by the Riemannian metric (i.e., drag coefficients), with unmodeled effects treated as noise in the least-squares estimation. More broadly, this framework can be extended to incorporate additional effects, provided they are observable in the system dynamics, as discussed later in this section.

Compared to a common data-driven alternative in geometric mechanics---directly learning a local connection that maps control inputs to passive-state motion \cite{bittner2018geometrically}---our approach is substantially more data efficient. By identifying a metric that preserves the structure of the dynamics, the resulting model generalizes across the configuration space, rather than being restricted to the shapes and trajectories observed during training.

In this section, we will use LandSalp as an example to demonstrate the data-driven geometric model identification method.

\subsection{Data Collection}

For data collection, we used the OptiTrack motion capture system to record the motion of each link of the LandSalp,\footnote{In this paper, we use motion capture to record the robot's motion relative to the world for a more intuitive demonstration. However, it is important to emphasize that the proposed method is not limited to this. If a state estimator can be implemented using onboard inertial measurements \cite{rollinson2011state, andrews2025modeling, andrews2025pose}, our method is capable of learning coefficients exclusively from proprioceptive data.} from which the system's position and joint angles were calculated. System velocities were obtained by applying Lie-group time differentiation to the position data in order to preserve group structure during differentiation.\footnote{Lie-group time differentiation is analogous to standard time differentiation, but it computes derivatives with respect to group actions rather than changes in coordinate values. For further details, please refer to Appendix~\ref{appendix:group}.} The driving torque was obtained from the torque sensing of the series-elastic actuator. The motion capture data was sampled at a frequency of 240 Hz, and the series-elastic actuator data was sampled at 200 Hz. All measurements were resampled to 200 Hz for subsequent unified processing. Additionally, all collected data were low-pass filtered to reduce noise.

When conducting the data collection experiments, we alternated the driving direction of each omniwheel and exhausted all possible driving combinations of the three wheels, so that the collected data could fully cover all possible couplings between the links. We recorded 8 runs in different orders and directions, making up a total of about 3 minutes of data. The input command was a constant frequency sine wave of 1/6 Hz, which is designed based on the actuator's maximum velocity and the omniwheel's maximum traction on the flat surface. Considering the proposed viscous drag-based model is first-order, the cutoff frequency of the low-pass filter was set according to the command frequency.

\subsection{Data-Driven Local Metrics}

As demonstrated in~\eqref{eq:pullback}, the actuation and viscous drag on each unit and joint can be pulled back to the configuration coordinates. Notably, the total configuration forces and moments are linear functions of the aggregated local metric~$\Localmetrics$
\begin{subequations} \label{eq:force}
\begin{align}
    \force_{\bundle} &= \transpose{\jac_{\bundle}}\Localmetrics\begin{bmatrix}
        \jac_{\bundle} & \jac_{\ctrl}
    \end{bmatrix}\begin{bmatrix}
        \bundledot \\
        \ctrl
    \end{bmatrix} \\
    &= \underbrace{\left(\transpose{\left(\begin{bmatrix}
        \jac_{\bundle} & \jac_{\ctrl}
    \end{bmatrix}\begin{bmatrix}
        \bundledot \\
        \ctrl
    \end{bmatrix}\right)}\kroneckerprod\transpose{\jac_{\bundle}}\right)}_{\jac_{\force_{\bundle}}}\vectorize\left(\Localmetrics\right),
\end{align}
\end{subequations}
where $\kroneckerprod$ is the Kronecker product and $\vectorize(\cdot)$ is the vectorization operation \cite{magnus2019matrix}. Similarly, by rewriting~\eqref{eq:forceconvert}, the force and moment acting on each individual salp unit, as well as its component in the direction of the series-elastic actuator, can be expressed as linear functions of the local metric, $\Localmetrics$:
\begin{eqalign}
    \selection\begin{bmatrix}
        \coforce_{1} \\ \vdots \\ \coforce_{\fstqty}
    \end{bmatrix} &= \selection\Localmetrics\begin{bmatrix}
        \jac_{\bundle} & \jac_{\ctrl}
    \end{bmatrix}\begin{bmatrix}
        \bundledot \\
        \ctrl
    \end{bmatrix} \\
    &= \underbrace{\left(\transpose{\left(\begin{bmatrix}
        \jac_{\bundle} & \jac_{\ctrl}
    \end{bmatrix}\begin{bmatrix}
        \bundledot \\
        \ctrl
    \end{bmatrix}\right)}\kroneckerprod\selection\right)}_{\jac_{\coforce}}\vectorize\left(\Localmetrics\right),
\end{eqalign}
where $\selection$ is the selection matrix that extracts the force component aligned with the series-elastic actuator.

Given the quasi-static equilibrium of the viscous drag-dominated system, as in~\eqref{eq:min_energy}, the total force and moment on the system are balanced, i.e. $\force_{\bundle} = 0$. Furthermore, the total net power on the system is also zero, i.e. $\left\langle \force_{\bundle}; \bundledot \right\rangle = 0$, providing a constraint that captures the dissipative structure of the low Reynolds number model and helps regularize identification of relative drag ratios among components. In addition, as shown in~\eqref{eq:forceconvert}, the force applied to each salp unit, $\coforce_{\fstidx}$, in the direction aligned with the series-elastic actuator, should correspond to the torque sensing of the series-elastic actuator, anchoring the absolute scale of the identified coefficients. These relationships enable an efficient data-driven modeling approach by leveraging motion capture data and the torque sensing of the series-elastic actuator. 

Specifically, the data-driven geometric model identification can be formulated as a semidefinite program that minimizes the difference between the model-predicted forces and the experimental data for a given motion, while regularizing the model using the zero total net power condition from the low Reynolds number swimming regime
\begin{eqalign}
 \min_{\Localmetrics} \quad& \left\|\jac_{\coforce}\vectorize\left(\Localmetrics\right) - \transpose{\begin{bmatrix}
        \tildecoforce_{1} & \cdots & \tildecoforce_{\fstqty}
    \end{bmatrix}}\transpose{\selection}\right\|^{2}_{2} \nonumber \\
    \quad&+ \lambda\left\langle \jac_{\force_{\bundle}}\vectorize\left(\Localmetrics\right), \bundledot \right\rangle \\
 s.t. \quad& \Localmetrics \succeq 0 \\
 & \Localmetrics \text{ is a diagonal matrix}
\end{eqalign}
where $\lambda$ is the strength of the regularization, and $\tildecoforce_{\fstidx}$ denotes the force measurement from the series-elastic actuator. The objective function is designed to learn the drag coefficient from torque sensing while ensuring that the system dynamics satisfy the low Reynolds number swimming condition. The first constraint guarantees that the local metric tensor is positive definite, and the second constraint ensures our assumptions that drag occurs only on each wheel and joint independently and that the drag on the wheel is aligned with its principal axis.\footnote{Modeling only the diagonal terms makes the model more robust and less prone to overfitting to limited data, but more coupled effects can be modeled if necessary when the rank (i.e., observability) of the data matrix is sufficient.} In practice, we follow the approach in~\cite{lee2020geometric} to make the problem coordinate-independent by first solving the unregularized problem and then using the error covariance to rescale the objective function. The resulting semidefinite program can be solved efficiently using standard convex-optimization tools; we use MOSEK for all results reported in this paper.

\begin{figure}[!t]
\centering
\includegraphics[width=\linewidth]{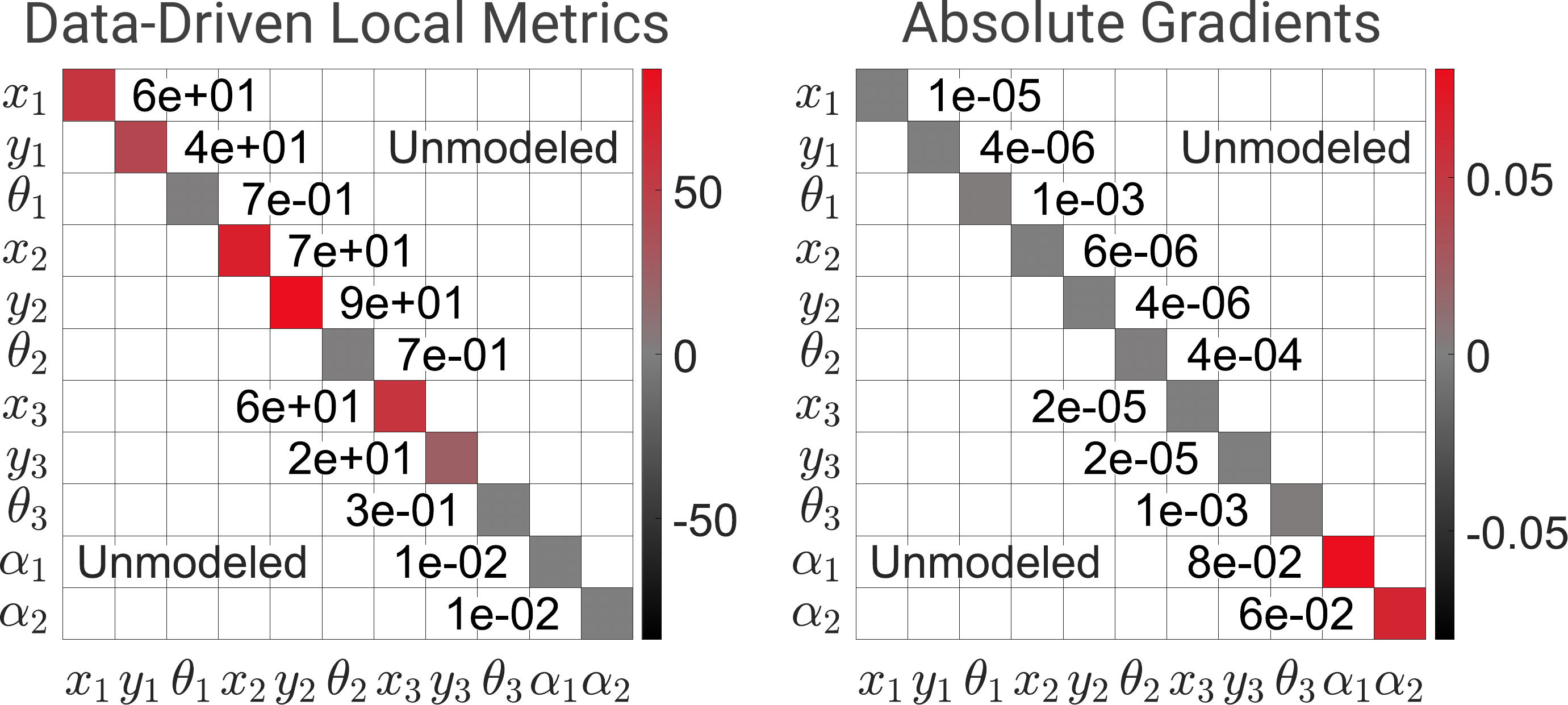}
\caption{Data-driven local metrics and corresponding residual gradients. In this paper, we assume that the drag of the wheels and joints is independent, and the principal axes of the local metric are aligned with the wheels. The white grids are unmodeled entries.}
\label{fig:localmetrics}
\end{figure}

Fig.~\ref{fig:localmetrics} shows the data-driven local metrics learned from experimental data and the corresponding residual gradients for each entry. The results are consistent with our intuition that the main driving force comes from drag in the direction of wheel rotation. Furthermore, the friction model accounts for extra viscous effects arising from the rolling of the omniwheel’s rollers, the steering of the wheels, the rolling resistance of the ball casters, and the rotation of the joints. It is important to note that the drag coefficients here are specific to their associated velocities, resulting in different units and scales when comparing linear drag to rotational drag. 

The large drag observed on the y-axis of the center wheel is likely due to an unmodeled inertia effect, specifically the force necessary to open or close the two side links when moving in the rolling direction of the center wheel. Additional coulomb friction might also exist in the experiment but considered negligible given the omniwheels and ball casters have only rolling friction on all directions. These unmodeled effects can be treated as disturbances---as they also would be in underwater deployments---and therefore provide a useful testbed for evaluating robustness of the feedback control methods in Section~\ref{sec:motion_planning}. In future work, such effects can also be modeled explicitly by augmenting the metric and associated dynamics, and incorporated into the identification framework (e.g., by extending the model class as in \cite{justus2024optimal}).

The residual gradients are small overall, indicating the stability of the data-driven model. The relatively high sensitivity in the drag of the two joints and the rotational drag of the wheels is due to the differing scales and units compared to the linear drag on the wheels.

It is important to highlight that the data-driven geometric model identification method is not limited to systems dominated by viscous drag. In principle, any locomotion system that can be modeled via Riemannian metrics can be similarly identified, provided that sufficient sensor data is available. As envisioned in Section~\ref{sec:conclusion}, our methodology is generalizable to systems operating at intermediate Reynolds numbers, which requires consideration of both inertial and drag effects, as well as joint stiffness. This finding shows the potential of our method to generalize across a wide range of environments and locomotion scenarios.

\section{Geometric Motion Planning and Control}
\label{sec:motion_planning}

Once the model coefficients have been determined through data-driven geometric model identification, we can proceed to consider motion planning and control for the salp-inspired robot. However, this task is not straightforward for two reasons:
\begin{enumerate}
    \item The control vector field $\cvf$ indicates that it is impossible to directly and instantaneously control all passive states of the system through linear combinations of control inputs. This limited instantaneous control authority requires the use of oscillatory inputs, which aim to achieve control of all passive states of the system through periodic control inputs within specific actuation limits. 
    \item Unlike the systems in extensive prior work on undulatory locomotion that assume direct shape control \cite{ostrowski1996mechanics, mcisaac2003motion, hatton2013geometric, xie2018kinematic, ramasamy2019geometry, hatton2022geometry}, the control inputs the salp-inspired robot are not aligned with the shape axes. This coupling between position and shape requires additional stability considerations because the shapes, which have inherent limits and influence the control vector field, cannot be directly controlled. 
\end{enumerate}
To address these issues that distinguish this system from conventional bio-inspired robotic systems, in this section, we will explore the application of geometric mechanics, averaging theory, and Lie group theory for geometric motion planning and feedback control of LandSalp to achieve omnidirectional locomotion and maintain stability. 

Specifically, this section will focus on the gait analysis and optimization of LandSalp. Gaits, commonly observed in animals such as horses running, humans walking, and snakes swerving, not only elevate controllability analysis from an instantaneous to a periodic level, but also better represent the water loading and ejection processes of salps.

\subsection{Controllability and Gait Analysis}

Given that the control vector field for an $n$-link salp is a $(2+\fstqty)$-by-$\fstqty$ matrix ($5$-by-$3$ for LandSalp), it is not feasible to achieve direct control over all passive components of the system instantaneously through linear combinations of control inputs. Here, we will analyze the system's controllability by extending the control authorities using first-order Lie brackets for oscillatory control inputs.

\begin{figure}[!t]
\centering
\includegraphics[width=\linewidth]{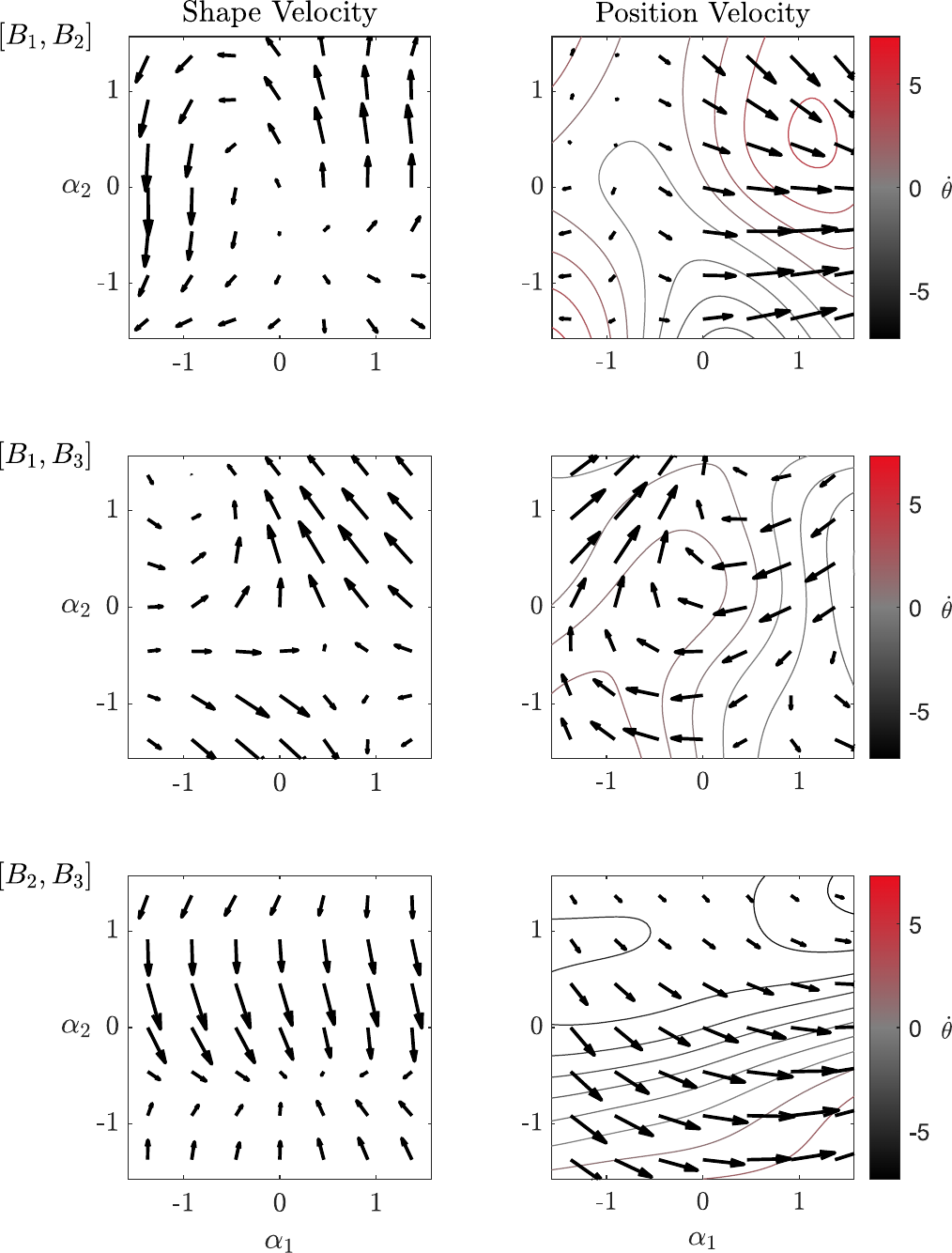}
\caption{First-order Lie brackets of the control vector fields for the LandSalp data-driven model. The left column shows the shape velocity results for each Lie bracket pair at different shapes, and the right column shows the position velocity results for each Lie bracket pair at different shapes, with the linear velocity in the $\dot{x}$ and $\dot{y}$ directions overlaid with the shape axes, and the contours showing the angular velocity $\dot{\theta}$.}
\label{fig:cvf_aug}
\end{figure}

Given the limited range of joint motion and the maximum velocity of the actuators, we use a periodic gait as the control input. A gait can be parameterized as a Fourier series\footnote{Physically, the periodic components in this context can be interpreted as cyclic variations in flow or force in the water jet cycle, and the constant component corresponds to the average flow through the jet.} consisting of a constant component $\bar{\ctrl}$ and $\trdidx$-th order oscillatory components determined by amplitudes $A^{\sin, \trdidx}$ and $A^{\cos, \trdidx}$ and a baseline frequency $\omega$\footnote{Taking into account the actuator's maximum speed limit, the wheel's maximum traction on a flat surface, and the approximate accuracy of truncation at the first-order Lie bracket, we have manually selected a control frequency as 1/6 Hz in this paper.}
\beq
 \ctrl(t) = \bar{\ctrl} + \sum^{\infty}_{\trdidx=1}\left(A^{\sin, \trdidx}\sin(2\trdidx\pi\omega t) + A^{\cos, \trdidx}\cos(2\trdidx\pi\omega t)\right).
\eeq
The LandSalp locomotion resulting from executing this gait around a nominal shape $\base^{\mathrm{nominal}}$ during a gait cycle can be expressed as a Baker-Campbell-Hausdorff (BCH) series and averaged using averaging theory. In this formulation, the geometrically averaged velocity $\bundledot^{\mathrm{avg}} \in \mathfrak{se}(2)\times\mathbb{R}^\sndqty$ (the constant body and shape velocity that produces the net displacement observed over the gait cycle) is obtained by taking the logarithm of the net displacement, $\bundle(T) \in \mathrm{SE}(2)\times\mathbb{R}^\sndqty$, divided by the cycle period, $T = 1/\omega$. The net displacement is represented as the continuous product of the exponentials corresponding to all of the input responses:
\begin{eqalign}
    \bundledot^{\mathrm{avg}} &= \frac{1}{T}\log{\bundle(T)} \\
    &= \frac{1}{T}\log\left(\Prodi^{T}_{0}\exp\left(\cvf\ctrl(t) dt\right)\right) \label{eq:prodint} \\
    &= \frac{1}{T}\int^{T}_{0}\cvf\ctrl(t) dt \nonumber\\
    &\quad+ \frac{1}{T}\int^{T}_{0}\int^{T}_{t}\frac{1}{2}[\cvf\ctrl(t), \cvf\ctrl(\tau)]d\tau dt + \cdots \label{eq:bch}\\
    &\approx \frac{1}{T}\int^{T}_{0}\cvf\ctrl(t) dt \nonumber\\
    &\quad+ \frac{1}{T}\int^{T}_{0}\int^{T}_{t}\frac{1}{2}[\cvf\ctrl(t), \cvf\ctrl(\tau)]d\tau dt \label{eq:trunbch}\\
    &= \cvf\bar{\ctrl} + \sum^{\sndqty}_{\fstidx=1}\sum^{\sndqty+1}_{\sndidx=\fstidx+1}\frac{1}{2\pi\omega}\ctrl_{[\fstidx, \sndidx]}[\cvf_{\fstidx}, \cvf_{\sndidx}], \label{eq:liebracket}
\end{eqalign}
where we first express~\eqref{eq:prodint} as a BCH series as demonstrated in~\eqref{eq:bch} \cite{radford1998local, magnus1954exponential}, then truncate it at the first-order Lie bracket as shown in~\eqref{eq:trunbch}, and finally integrate the controlling Fourier series form as presented in~\eqref{eq:liebracket}. $[\cvf_{\fstidx}, \cvf_{\sndidx}]$ represents the Lie bracket of the control vector fields $\cvf_{\fstidx}$ and $\cvf_{\sndidx}$, and $\ctrl_{[\fstidx, \sndidx]}$ denotes the corresponding coefficient in the Fourier series representation of the periodic control input:
\begin{eqalign}
    \ctrl_{[\fstidx, \sndidx]} &= \sum^{\infty}_{\trdidx=1}\frac{1}{k}\bigg(\bar{\ctrl}_{\sndidx}A^{\sin, \trdidx}_{\fstidx} - \bar{\ctrl}_{\fstidx}A^{\sin, \trdidx}_{\sndidx} \nonumber \\
    &\quad \left. - \frac{1}{2}A^{\sin, \trdidx}_{\fstidx}A^{\cos, \trdidx}_{\sndidx} + \frac{1}{2}A^{\cos, \trdidx}_{\fstidx}A^{\sin, \trdidx}_{\sndidx}\right) \\
    [\cvf_{\fstidx}, \cvf_{\sndidx}] &= \parderiv{\cvf_{\sndidx}}{\bundle}\cvf_{\fstidx} - \parderiv{\cvf_{\fstidx}}{\bundle}\cvf_{\sndidx} \\
    &= \begin{bmatrix}\parderiv{\cvf_{\sndidx}^{\fiber}}{\fiber} & \parderiv{\cvf_{\sndidx}^{\fiber}}{\base} \\ \parderiv{\cvf_{\sndidx}^{\base}}{\fiber} & \parderiv{\cvf_{\sndidx}^{\base}}{\base}\end{bmatrix}\begin{bmatrix}\cvf_{\fstidx}^{\fiber} \\ \cvf_{\fstidx}^{\base}\end{bmatrix} - \begin{bmatrix}\parderiv{\cvf_{\fstidx}^{\fiber}}{\fiber} & \parderiv{\cvf_{\fstidx}^{\fiber}}{\base} \\ \parderiv{\cvf_{\fstidx}^{\base}}{\fiber} & \parderiv{\cvf_{\fstidx}^{\base}}{\base}\end{bmatrix}\begin{bmatrix}\cvf_{\sndidx}^{\fiber} \\ \cvf_{\sndidx}^{\base}\end{bmatrix} \\
    &= \begin{bmatrix}\left[\cvf_{\fstidx}^{\fiber}, \cvf_{\sndidx}^{\fiber}\right] + \left(\parderiv{\cvf_{\sndidx}^{\fiber}}{\base}\cvf_{\fstidx}^{\base} - \parderiv{\cvf_{\fstidx}^{\fiber}}{\base}\cvf_{\sndidx}^{\base}\right) \\ \left(\parderiv{\cvf_{\sndidx}^{\base}}{\base}\cvf_{\fstidx}^{\base} - \parderiv{\cvf_{\fstidx}^{\base}}{\base}\cvf_{\sndidx}^{\base}\right)\end{bmatrix},
\end{eqalign}
where all control vector fields $\cvf$ and their derivatives are evaluated at the nominal shape, $\cvf^{\fiber}$ and $\cvf^{\base}$ are the rows corresponding to the body and shape velocities, and $\cvf_{\fstidx}$ and $\cvf_{\sndidx}$ are the columns corresponding to the $\fstidx$- and $\sndidx$-th control inputs. 

The accuracy of the approximation by truncation depends on three key factors: the frequency $\omega$, the magnitude of the control vector field $\cvf$ in the control direction (which is influenced by the choice of coordinates describing the multi-link system's overall motion), and the amplitude of the control input. This dependency arises because all higher-order terms in the BCH series are products of exponential functions of the period $T$, high-order Lie brackets of the control vector field $\cvf$, and multinomial terms of the mean and variation of the control input. A more detailed discussion on the approximation accuracy and the choice of coordinates to optimize this accuracy is provided in Appendix~\ref{appendix:coordinate}. The aforementioned methodology can also be employed to determine the locomotion displacement resulting from periodic shape control, yielding results that align with prior research findings \cite{hatton2015nonconservativity, bass2022characterizing}, as discussed in Appendix~\ref{appendix:shape}.

The first-order Lie brackets constitute an additional set of vector fields, as shown in Fig.~\ref{fig:cvf_aug}. These vector fields represent the passive state velocities that arise from alternating pairs of control inputs over a cycle. These additional vector fields complement the original control vector fields for states that cannot be directly and instantaneously controlled. By compiling all pairs of Lie brackets along with their corresponding amplitudes and appending them to the original control vector, we construct the augmented control vector field and augmented control input
\begin{eqalign}
 \cvf^{\mathrm{aug}} &= \begin{bmatrix}\cvf & [\cvf_{\fstidx}, \cvf_{\sndidx}] \end{bmatrix} \nonumber\\ 
 &\quad \mathrm{for}\ \fstidx = 1,\cdots,\fstqty-1,\ \sndidx = \fstidx+1,\cdots,\fstqty \\
 \ctrl^{\mathrm{aug}} &= \begin{bmatrix}\bar{\ctrl} \\ \frac{1}{2\pi\omega}\ctrl_{[\fstidx, \sndidx]}\end{bmatrix} \nonumber\\ 
 &\quad \mathrm{for}\ \fstidx = 1,\cdots,\fstqty-1,\ \sndidx = \fstidx+1,\cdots,\fstqty.
\end{eqalign}
Combining the first-order Lie bracket with the original control vector field, the augmented control vector field satisfies the Lie algebra rank condition and ensures the small-time local controllability of the system.\footnote{This nonlinear controllability statement is local: because the averaged dynamics depend on shape, controllability is guaranteed only in a neighborhood of the nominal shape. Transitions between distant shapes generally require sequencing gaits about multiple nominal shapes.}

\subsection{Motion Planning and Feedback Control}

\begin{figure}[!t]
\centering
\includegraphics[width=\linewidth]{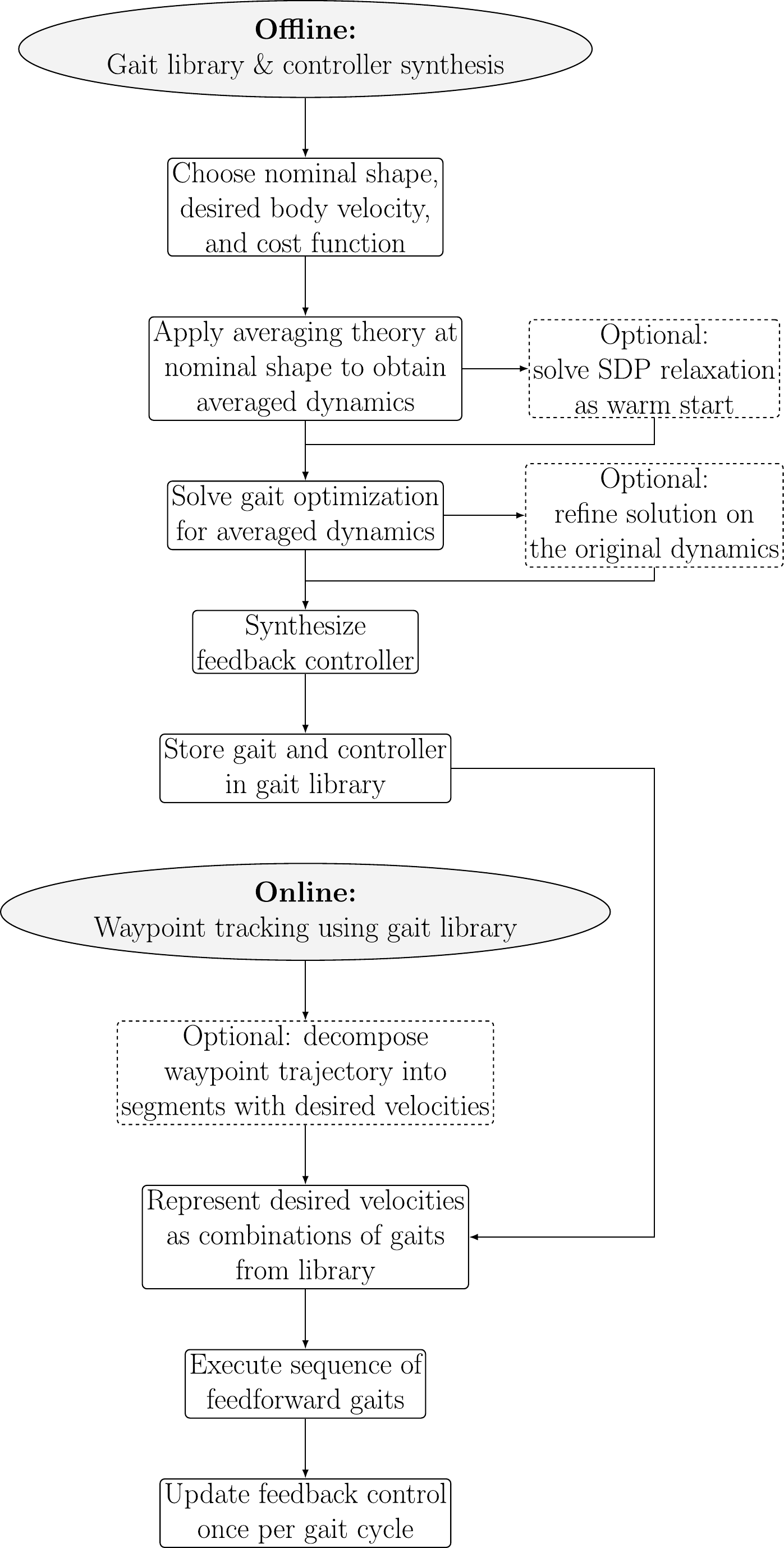}
\caption{Flowchart of the proposed motion planning and control methodology, comprising offline gait and control synthesis and online execution.}
\label{fig:flowchart}
\end{figure}

Figure~\ref{fig:flowchart} summarizes the proposed approach. Offline, we construct a library of gaits and associated feedback controllers. Online, a desired waypoint trajectory can be decomposed into segments with associated desired velocities. These velocities are represented as combinations of gaits from the library, which are executed sequentially. The corresponding feedback controller is updated once per gait cycle as the phase wraps around.

Specifically, based on the averaging results, given the desired average velocity $\bundledot^{\mathrm{desired}}$, the control parameters in the form of a $\trdqty$-order Fourier series $\Ctrl = \transpose{\begin{bmatrix}\bar{\ctrl} & A^{\sin, 1} & A^{\cos, 1} & \cdots & A^{\sin, \trdqty} & A^{\cos, \trdqty}\end{bmatrix}}$ around the nominal shape $\base^{\mathrm{nominal}}$ can be determined by formulating a nonlinear programming problem
\begin{eqalign}
    \min_{\Ctrl} \quad& \|\Ctrl\|^2_{\costmetric} \label{eq:cost} \\
    s.t. \quad& \bundledot^{\mathrm{avg}} = \bundledot^{\mathrm{desired}}, \label{eq:velocity}
\end{eqalign}
where the control cost metric $\costmetric$ can be selected to regularize (a) the average squared jet velocities, (b) the average dissipated power, or (c) the average squared thrust force:
\begin{eqalign}
    \|\Ctrl\|^2_{\costmetric} &= \frac{1}{T}\int_{0}^{T}\|\ctrl(t)\|^2_{\costmetric_{\ctrl}} dt \\
    \costmetric_{\ctrl} &= 
    \begin{cases}
    \diag\left(\frac{1}{\ctrl_{1, \mathrm{max}}^2}, \frac{1}{\ctrl_{2, \mathrm{max}}^2}, \frac{1}{\ctrl_{3, \mathrm{max}}^2}\right) & \mathrm{(Jet\ Velocity)} \\
    \transpose{\jac}_{\ctrl}\Localmetrics\jac_{\bundle}\cvf & \mathrm{(Power)} \\
    \transpose{\jac}_{\ctrl}\transpose{\Localmetrics}\Localmetrics\jac_{\ctrl} & \mathrm{(Force)},
    \end{cases} \\
    \costmetric &= \diag\left(\costmetric_{\ctrl}, \frac{1}{2}\costmetric_{\ctrl}, \frac{1}{2}\costmetric_{\ctrl}, \dots, \frac{1}{2}\costmetric_{\ctrl}, \frac{1}{2}\costmetric_{\ctrl}\right),
\end{eqalign}
where $\ctrl_{\fstidx, \mathrm{max}}$ is the maximum jet velocity (equivalent to the maximum velocity of the actuator in our LandSalp robot) for the $\fstidx$-th actuator, and $\costmetric_{\ctrl}$ is a cometric tensor representing the cost of instantaneous control, which can be proposed in terms of jet velocity, power, or torque regularization.

\begin{figure}[!t]
\centering
\includegraphics[width=\linewidth]{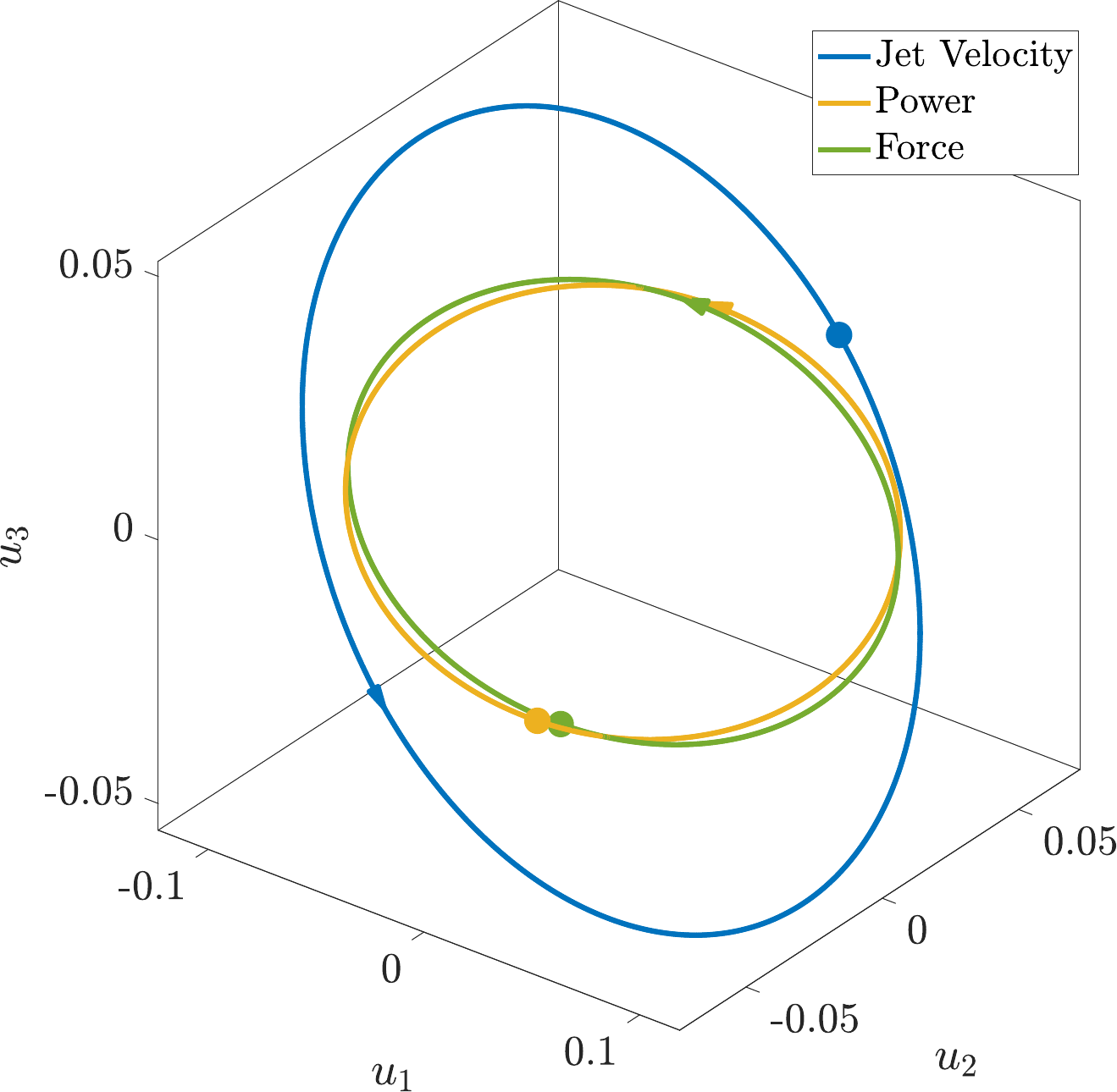}
\caption{Optimal forward gaits corresponding to three different cometrics at the $(0, 0)$ shape, where circle and triangle markers denote the initial phase and the gait direction, respectively. The velocity metric focuses on penalizing the speed of the second wheel, resulting in a contraction along that direction and an expansion in others. The power metric has a more uniform penalty across all three wheels, yielding a gait resembling a circle. Lastly, the torque metric imposes greater penalties on the first and second wheels compared to the power metric, consequently resulting in increased speeds for the third wheel.}
\label{fig:gait_metrics}
\end{figure}

Fig.~\ref{fig:gait_metrics} shows the forward gait corresponding to these three different cometrics in a straight configuration (zero joint angles). Given that the second actuator has a smaller maximum speed limit, jet velocity regularization is flatter along the $u_2$ direction, resulting in a slower gait cycle in $u_2$. Conversely, to minimize the power and forces that are neutralized within the system, both power and force regularizations ensure that the three wheels operate at similar speeds, because they are nearly perpendicular to each other in a straight configuration. 

\begin{figure*}[!t]
\centering
\includegraphics[width=\linewidth]{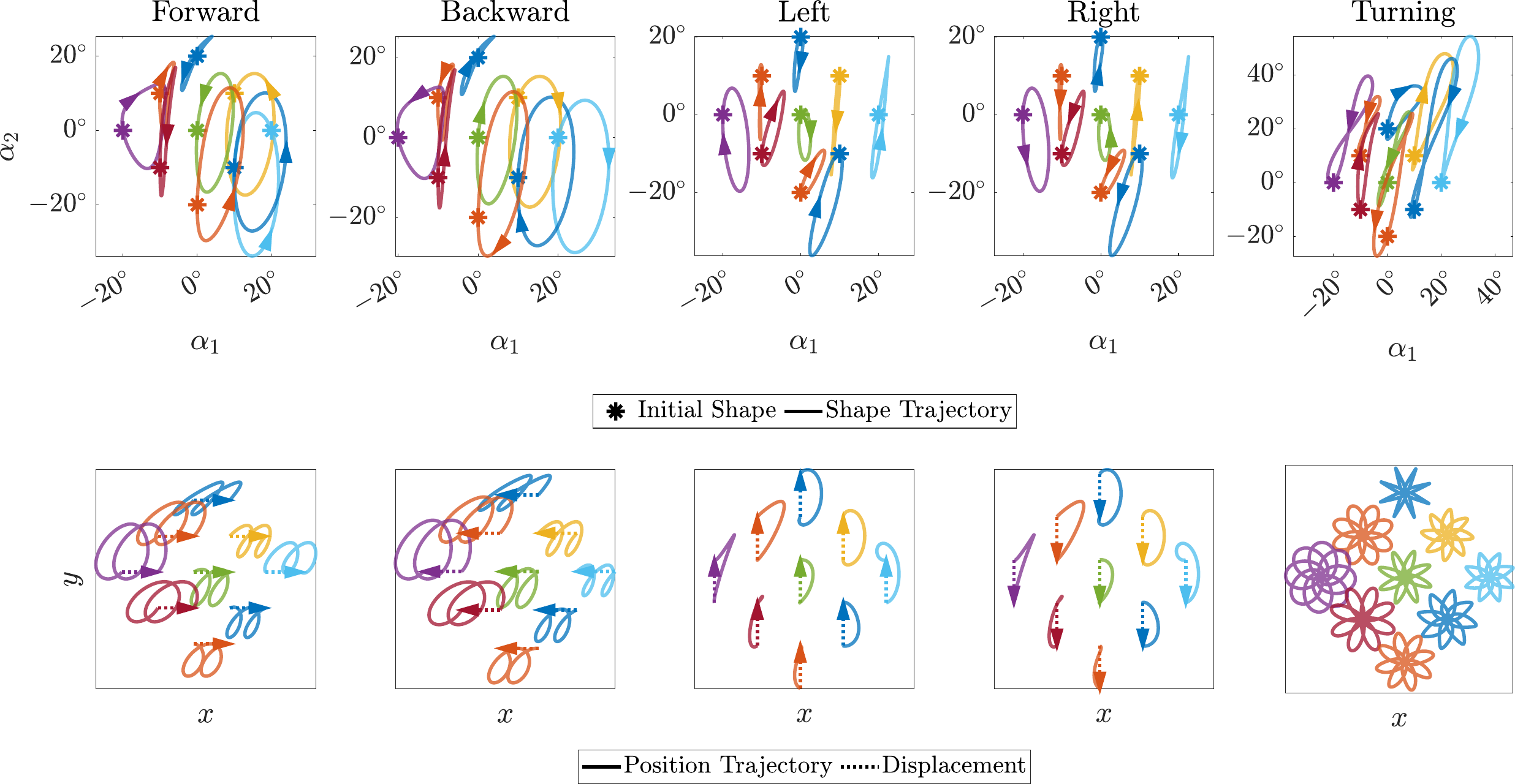}
\caption{LandSalp simulation results for forward, backward, left, right, and turning motions. The first row shows the shape trajectory and nominal shape of the robot, with arrows indicating the direction of motion. The second row illustrates the position trajectory and net displacement of each nominal shape over 2, 1, and 8 gait cycles corresponding to forward-backward, left-right, and turning motions, respectively.}
\label{fig:simulation_results}
\end{figure*}

For the forward gait, the shape changes only slightly during the period (approximately 10 degrees in the first joint), as shown in Fig.~\ref{fig:simulation_results}. Consequently, the power regularization remains nearly constant and is similar to the force regularization. Compared to power regularization, force regularization causes the first and third wheels to drive slightly faster while slowing down the second wheel. This difference arises because force regularization incorporates the square of the drag coefficients when calculating the required cost, penalizing the second wheel slightly more. These regularized metric tensors are particularly useful when users aim to minimize specific costs based on the capabilities of the actuators.

The primary source of nonlinearity in this problem is the nonlinear constraint~\eqref{eq:velocity}, which poses significant challenges in optimization. Utilizing the analysis based on averaging theory as discussed in~\eqref{eq:liebracket}, we can approximate this constraint by the averaged results. This approach allows us to derive analytical gradients and Hessians from the analysis, achieving a more tractable optimization process.
\begin{eqalign}
    \|\Ctrl\|^2_{\costmetric} \quad& \transpose{\Ctrl}\costmetric\Ctrl \label{eq:approxcost} \\
    s.t. \quad& \cvf^{\mathrm{aug}}\ctrl^{\mathrm{aug}}(\Ctrl) = \bundledot^{\mathrm{desired}}.
\end{eqalign}
This gait optimization formulation on the averaged system pose a quadratic constraint quadratic programming problem; it can be relaxed to a semidefinite program (SDP) to obtain a computationally efficient warm start \cite{boyd2023convex}. While the relaxation need not be tight, it often provides a useful initialization for subsequent nonlinear programming to refine a locally optimal gait. 
Furthermore, after determining the optimal gait for the averaged system, it is optional to refine this gait in the full system using a finite difference method. If the feedback controller evaluates the error at a specific phase (as described below), rather than averaging the error over the entire gait cycle (as discussed in Appendix~\ref{appendix:average}), then ensuring that the shape trajectory passes through the nominal shape at that phase may require this refinement.

Because the optimal gait is computed around a nominal shape, any perturbation or model mismatch may cause deviation from this nominal shape, thus invalidating the optimal gait. Therefore, we incorporate feedback control into the system. Following the methodology proposed in~\cite{teel1992nonholonomic, vela2003averaging}, we linearize the averaged system at the nominal shape $\base^{\mathrm{nominal}}$ and the optimal gait $\Ctrl$
\begin{align}
    \Delta\bundledot &\approx \parderiv{\cvf^{\mathrm{aug}}(\base^{\mathrm{nominal}})\ctrl^{\mathrm{aug}}(\Ctrl)}{\bundle}\Delta\bundle \nonumber \\
    &\quad + \parderiv{\cvf^{\mathrm{aug}}(\base^{\mathrm{nominal}})\ctrl^{\mathrm{aug}}(\Ctrl)}{\Ctrl}\Delta\Ctrl,
\end{align}
and then discretize\footnote{In our implementation, we apply an exponential mapping to the logarithmic average velocity to obtain the discrete system dynamics.} by the gait period $T$ to design a discrete linear quadratic regulator with the state feedback gain $\feedback$ for updating the control $\Ctrl$ at each gait cycle to correct the error in $\bundle = \transpose{\begin{bmatrix}\transpose{\fiber} & \transpose{\base}\end{bmatrix}}$
\begin{align} \label{eq:feedback}
    &\Ctrl(t) = \Ctrl_0 - \feedback\left(\bundle^{\mathrm{desired}}(t_\trdidx) - \bundle(t_\trdidx)\right) \nonumber \\
    &\quad \text{for } t\in[t_\trdidx, t_{\trdidx+1}), \text{where } t_\trdidx=\trdidx T, \trdidx\in\mathbb{Z}_{\ge 0},
\end{align}
where $\Ctrl_0$ is the feedforward optimal gait. As discussed in~\cite{vela2003averaging}, because the averaged system satisfies the Lie algebra rank condition, the full system can be stabilized by stabilizing the averaged system.

Note that when the shape velocity is set to zero, this formulation guarantees that the initial and terminal shapes remain unchanged. In our implementation, we define the initial shape as the nominal shape $\base^{\mathrm{nominal}}$, which implies that the shape trajectory will pass through this nominal shape. However, this choice may not always be advantageous. For instance, if the nominal shape is close to the joint limits, the resulting trajectory may be more likely to violate these constraints. Moreover, the average shape over the gait cycle might not align with the nominal shape, thereby degrading the approximation accuracy of the averaging theory. In practical applications, the proposed feedback controller mitigates this issue and compensates for model mismatches and external disturbances. Readers interested in a more theoretical treatment of this problem are encouraged to read Appendix~\ref{appendix:average}, which provides additional discussion and alternative formulations to ensure that the shape trajectory over the gait cycle averages at the desired shape.

\section{Experiment Evaluation}

To assess the efficacy of the proposed modeling and control methodology, we conduct experiments in both simulation and hardware. In simulation, we employ the data-driven geometric model learned from experimental data to demonstrate the capability of our approach to finding optimal gaits for different nominal shapes and desired velocities. On the LandSalp hardware platform, we execute these optimal gaits and corresponding feedback controllers to validate their locomotion performance as well as measure displacement errors and evaluate the precision of the model predictions.

\subsection{Simulation Evaluation}

We first used the proposed motion planning method to generate optimal gaits for various locomotion directions and nominal shapes to test its effectiveness. We explored five different locomotion modes: forward, backward, left, right, and clockwise turning. The gait period was set at 1/6 Hz. The velocities for forward and backward locomotion were 3 cm/cycle, while for left and right motions, the velocities were 6 cm/cycle. The turning speed was 45 deg/cycle. These velocity parameters were designed to ensure that the maximum drive speed remained within the actuator limitations and maintained traction without slipping on a flat surface. All gaits were optimized to minimize the mechanical power of the actuators. In the evaluation, all gaits are characterized using a first-order Fourier series. The simulation model used physical coefficients derived from experimental data.

Fig.~\ref{fig:simulation_results} illustrates the shape and position trajectories for locomotion in various directions and for different nominal shapes. In these experiments, forward and backward motions are executed for 2 cycles, left and right motions for 1 cycle, and turning motions for 8 cycles to enhance visualization clarity. The results demonstrate that the proposed motion planning algorithm effectively finds the open-loop optimal gaits that maintain periodic equilibrium and drive the robot in the desired direction. 

Notably, turning or lateral movements exhibit greater efficiency compared to forward and backward movements. This increased efficiency can be attributed to the alternating lateral deflection of the wheels along the link's long axis, which leads to significant internal cancellation of driving forces during longitudinal motion, but much less cancellation during lateral and turning locomotion. The optimal gaits for locomotion in opposite directions are symmetrical and phase-opposite. However, the open-loop stability of gaits in opposite directions varies, leading to significant differences in feedback control, which will be further discussed in the hardware evaluation section.

The range of shape motion for some gaits is significantly larger than for others, depending on the nominal shape, which may increase the likelihood of failure and approaching joint limits when implemented on hardware. Turning motions, especially, cause substantial joint movements, particularly when the nominal shape is distant from the straight configuration.
Such large deviations from the nominal shape may invalidate the averaging around the nominal shape. To mitigate this, a potential solution is to explicitly enforce joint limits and constrain the average of the shape trajectories to match the nominal shape during optimization, thereby ensuring feasibility (see Appendix~\ref{appendix:average}). Overall, the nominal shape at the straight configuration has the smallest shape motion range and remains distant from the joint limits.

\subsection{Hardware Evaluation}

\begin{figure}[!t]
\centering
\includegraphics[width=\linewidth]{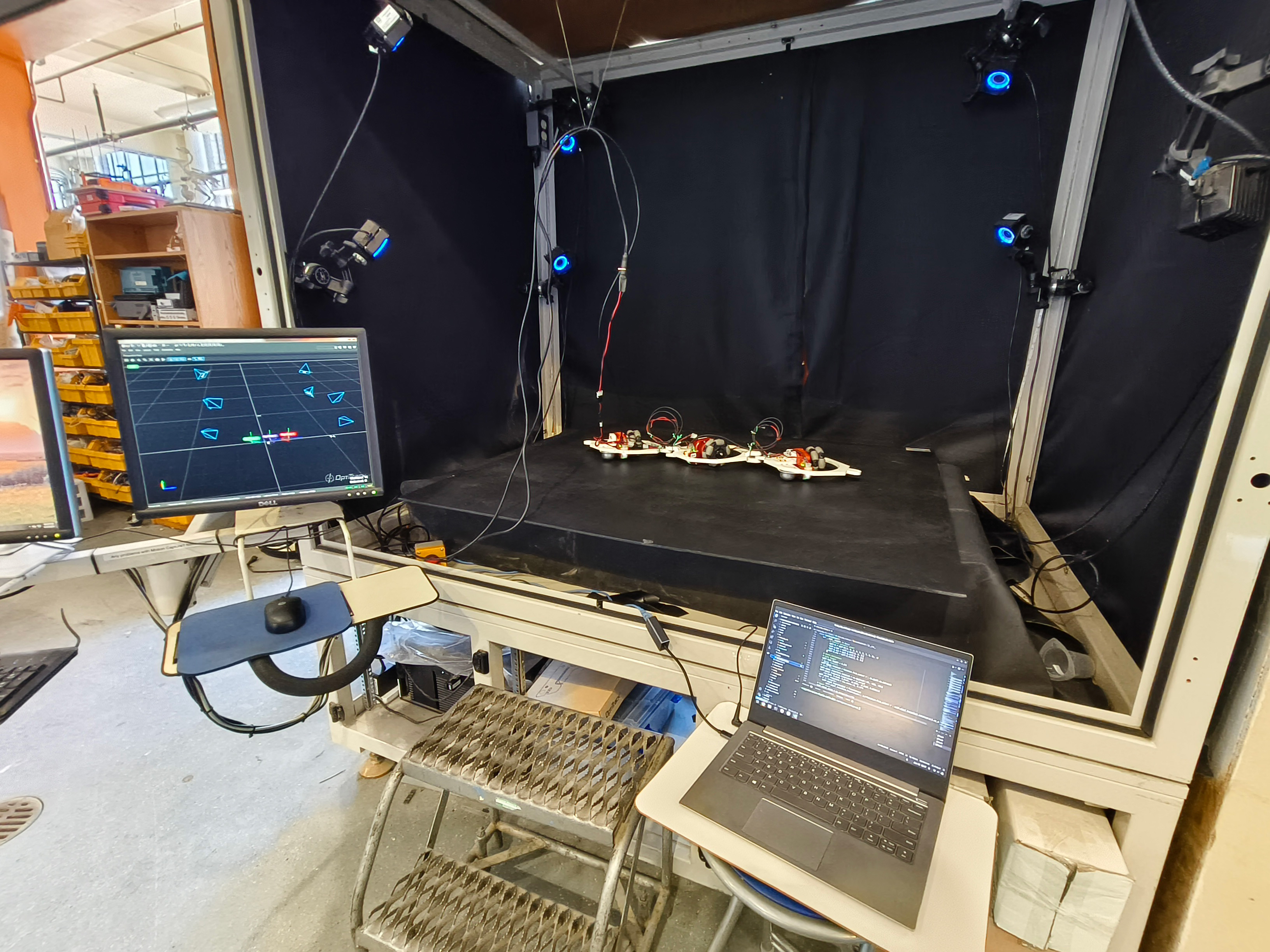}
\caption{LandSalp experiment setup. The robot runs on a platform integrated with the OptiTrack motion capture system. Motion capture data is processed by a desktop computer and subsequently transmitted via the Robot Operating System (ROS) to a laptop that executes the control algorithm. The robot is connected by two cables, one for power supply and the other for command signals.}
\label{fig:system_setup}
\end{figure}

We conducted the hardware evaluation of the proposed methods by commanding the LandSalp to move in different directions while maintaining a nominal shape of zero joint angles. The experimental setup is shown in Fig.~\ref{fig:system_setup}. The LandSalp robot ran on a flat platform, utilizing OptiTrack motion capture for state estimation, feedback control, and ground truth data collection. The robot was powered through tension-free cables and controlled via an external laptop connected through Ethernet. The laptop ran ROS Noetic, receiving motion capture data at a frequency of 240 Hz and transmitting control commands to each actuator at a frequency of 200 Hz.

Given that the bottleneck of the system is the maximum speed of the actuators, we have chosen to regularize the jet velocity within the motion planning and control processes. 
Furthermore, we implement the discrete feedback as proposed in~\eqref{eq:feedback} within each cycle.
In our implementation, all gaits are represented using a first-order Fourier series.

\begin{figure}[!t]
\centering
\includegraphics[width=\linewidth]{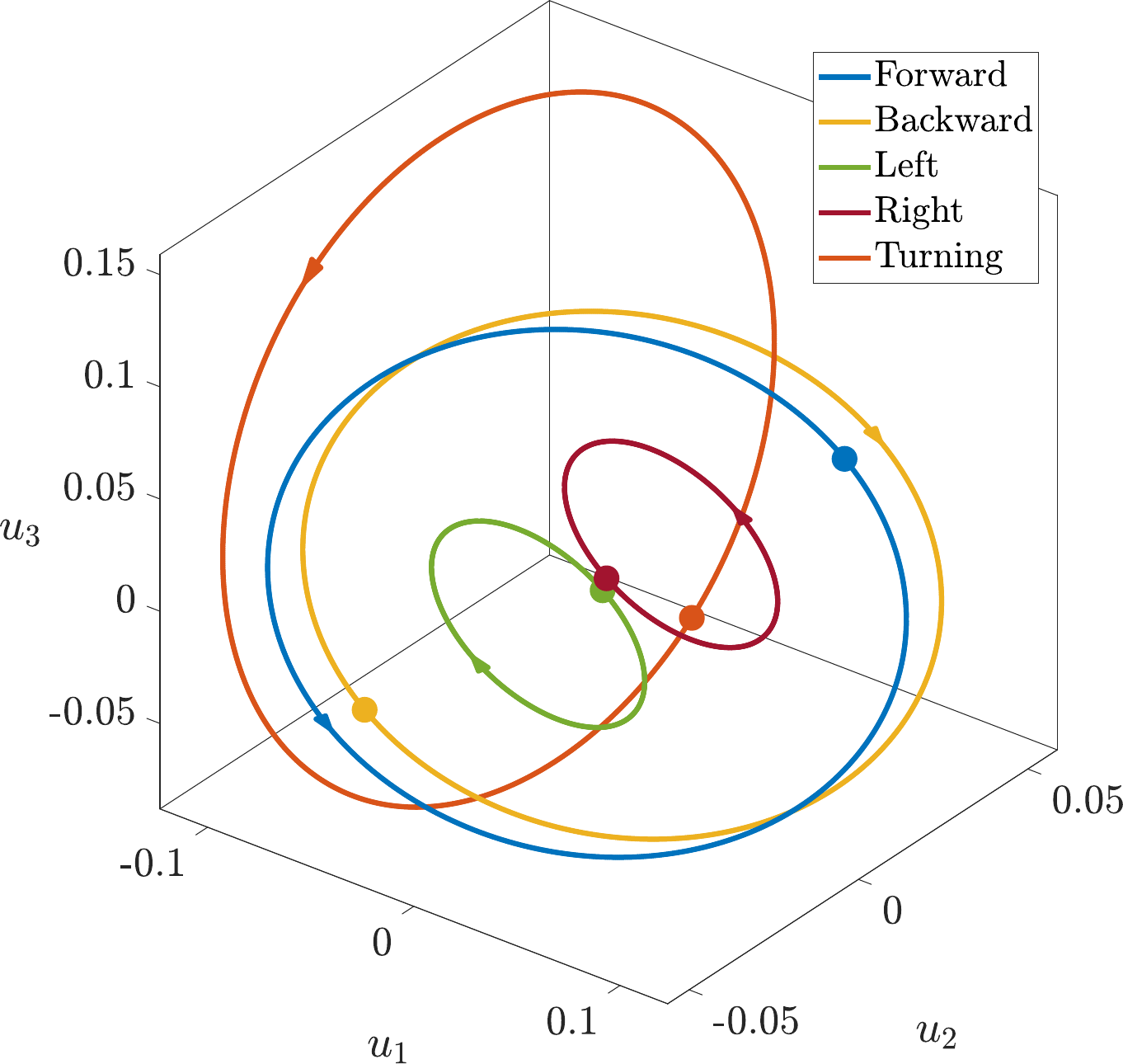}
\caption{Optimal gaits for forward, backward, left, right, and turning motions under the velocity cost. The circle and triangle markers denote the initial phase and the gait direction, respectively. A notable observation is that opposite motions have mirrored gaits relative to the zero joint angles, despite variations in linearization and feedback control.}
\label{fig:gait}
\end{figure}

\subsubsection{Omnidirectional Locomotion}

\begin{figure*}[!t]
\centering
\includegraphics[width=\linewidth]{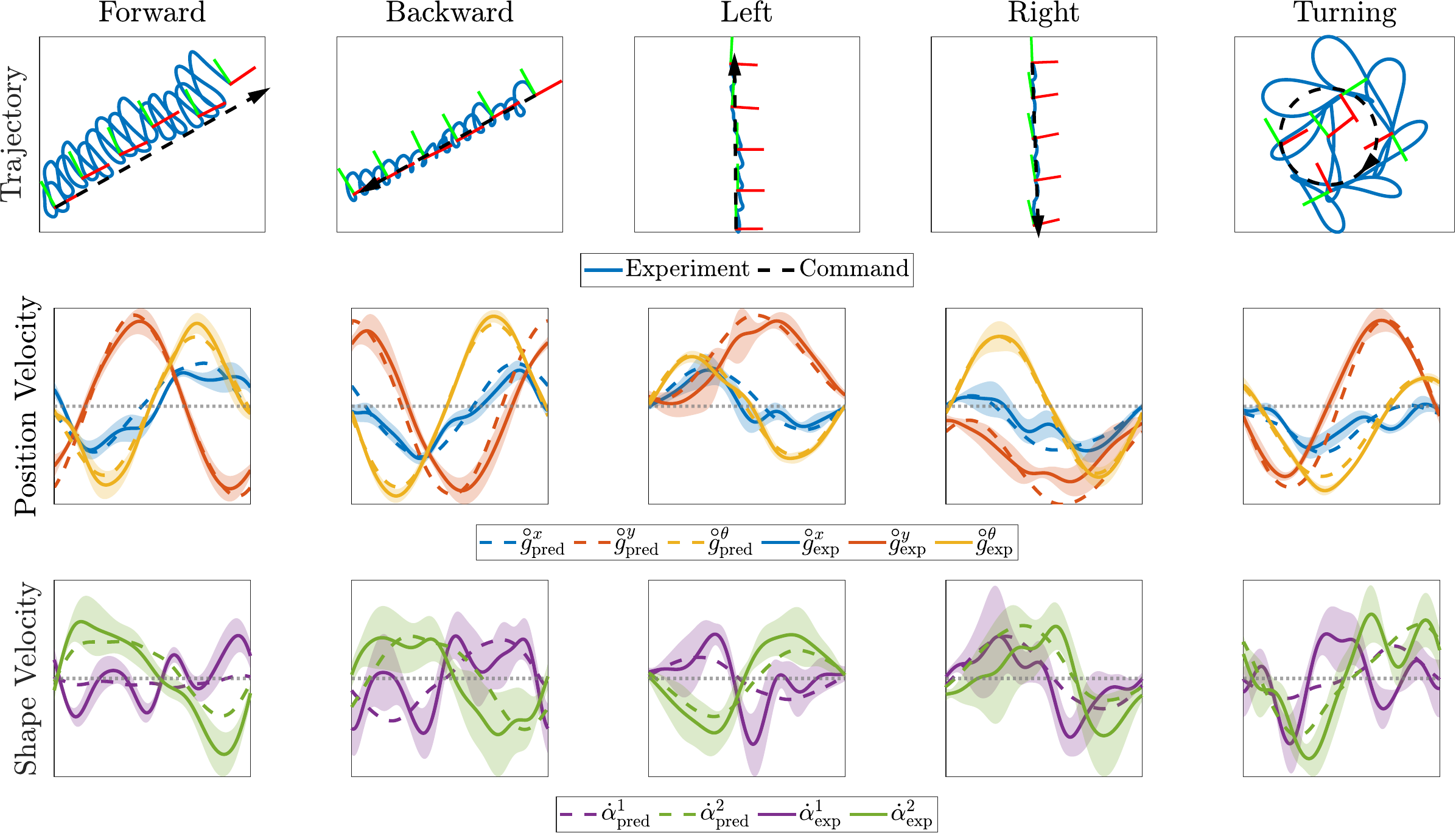}
\caption{Experimental results of LandSalp's forward, backward, left, right, and turning motions. The first row shows the robot's commands and the experimental trajectories and orientations. The second row illustrates the average model-predicted and experimental position velocities, as well as the corresponding standard deviations, over a gait cycle. The third row displays the average model-predicted and experimental shape velocities, along with their standard deviations, over a gait cycle.}
\label{fig:experiment_velocity}
\end{figure*}

Fig.~\ref{fig:gait} shows the optimal gaits tested in the hardware evaluation. Similar to the simulation evaluation, the robot was commanded to move forward, backward, left, right, and clockwise at its nominal shape with zero joint angles. The forward and backward motions were executed at a speed of 3 cm/cycle for 15 cycles, the left and right movements were conducted at a speed of 6 cm/cycle for 8 cycles, and the turning motion was performed at 45 deg/cycle for 8 cycles. The forward-backward and turning gaits are performed with actuator speeds close to maximum, while the lateral gaits are relatively more efficient. 

Consistent with theoretical predictions from simulations, the optimal gaits for two opposite directions were symmetrical within the control space and had reverse phases. However, their stability and feedback control varied significantly due to the asymmetry of the control vector field, which is also evident from the robot's asymmetric geometry. This asymmetry is the primary reason for excluding counterclockwise motion in the experiments with the used robot hardware configuration. Intuitively, it is evident from the robot's geometry that instability arises in the counterclockwise direction because the required inward driving force of the side wheels induces joint buckling and link collision. Conversely, clockwise motion is more easily stabilized as the outward movement of the side wheels generates the required torque while also maintaining the robot's shape by pulling it apart from both sides.

The first row of Fig.~\ref{fig:experiment_velocity} illustrates the locomotion trajectories for the tested motions. The frame and trajectory represent the system body frame, determined by the center of mass and the average orientation of the three links. The robot effectively tracks commands in all directions, although with greater accuracy in backward compared to forward motions, and leftward compared to rightward movements. This discrepancy can be attributed to the asymmetric geometry of the robot and the consequent asymmetric control vector field, as discussed previously. Consequently, the feedback controller requires more effort to stabilize motion in one direction compared to its opposite. Feedback that actively compensates for locomotion errors due to unmodeled effects or disturbances is crucial. Specifically, in forward motion, feedback control mitigates lateral drift, and during rightward motion, it minimizes orientation errors. In summary, the tracking error for all directions stays small enough to be compensated for by a higher-layer tracking controller.

\begin{figure*}[!t]
\centering
\includegraphics[width=\linewidth]{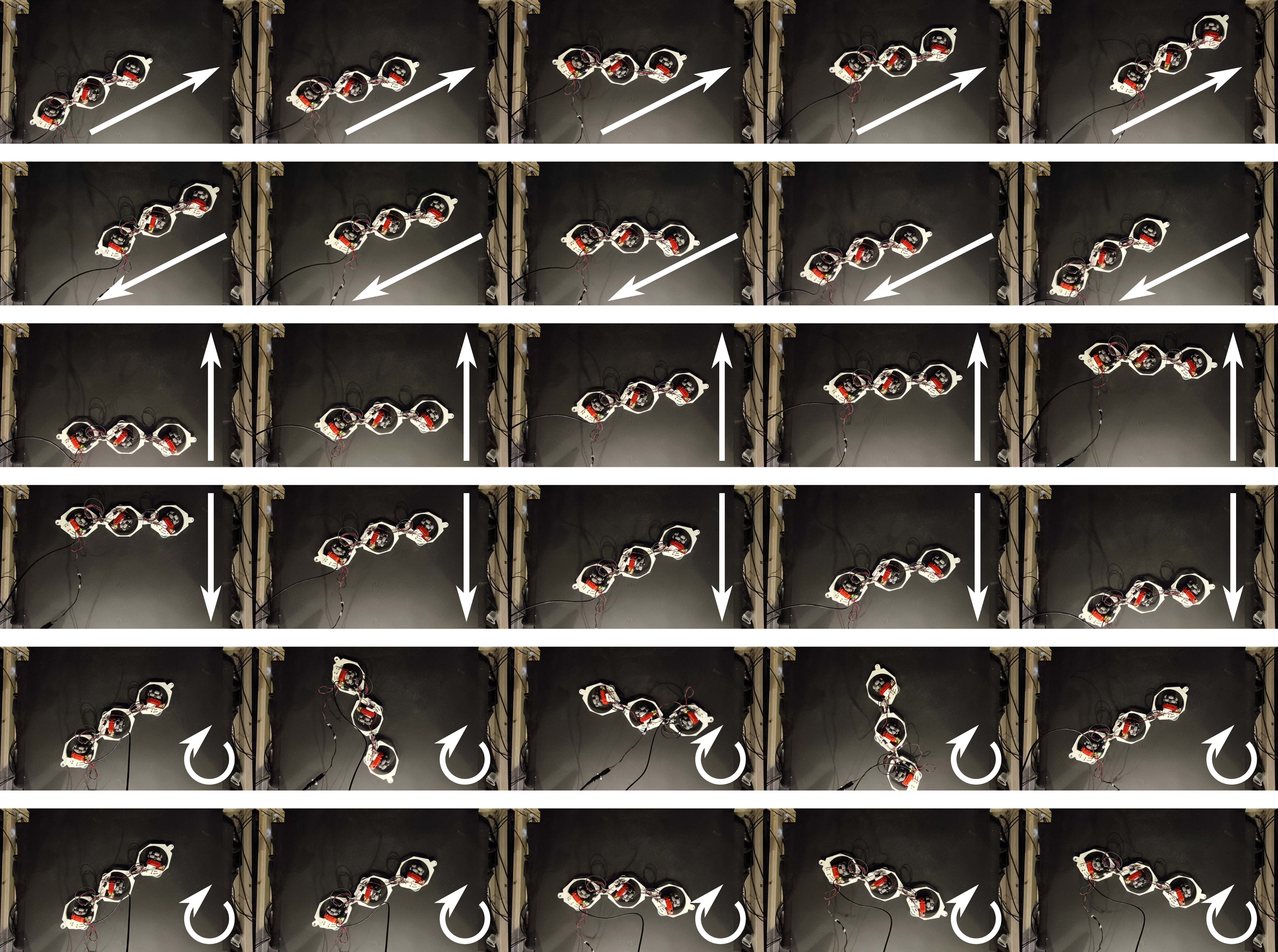}
\caption{Snapshots of the forward, backward, left, right, and turning motions of the LandSalp experiment. The first four rows show the robot's positions at approximately the quarter intervals of the entire motion cycle, as well as the configurations corresponding to the quarter intervals of the gait cycle. The last row presents the robot's configurations at the quarter points of the gait cycle during a single gait of clockwise turning. The white arrows indicate the desired direction of velocity.}
\label{fig:experiment_diagram}
\end{figure*}

The second and third rows of Fig.~\ref{fig:experiment_velocity} show the average values of both experimental and model-predicted position and shape velocities across all cycles of the experiment. The shaded regions represent the standard deviations of these statistical values. It is evident that the shape velocity varies more than the position velocity across different gait cycles. This increased variability may be attributed to the passive nature of the joints, which are more susceptible to perturbations than the overall system position. Additionally, the position velocity in lateral motion shows slightly more variation than in other directions, possibly due to the higher command velocity for lateral motion. More importantly, the model's predictions effectively capture the primary trends in both position and shape velocities, thereby ensuring good control performance. The model's predictions for position velocity are more accurate than those for shape velocity, where the experimental data indicate some second-order effects resembling spring behavior caused by the harness on the robot, which exceeds the scope of our first-order model. Nevertheless, the model successfully captures the first-order trends in joint velocities. Extending our model and methodology to incorporate second-order effects is a key element of our envisioned future work.

Each row of Fig.~\ref{fig:experiment_diagram} illustrates a series of snapshots depicting the LandSalp locomoting in various directions. From the snapshots, we can again find the symmetry between the two pairs of opposite locomotion directions: forward-to-backward and left-to-right. During forward or backward movement, LandSalp first bends its head and tail to align the omniwheels' rollers to minimize drag, then drives the robot forward using the central wheel, and finally returns the head and tail back. For lateral motions, the LandSalp first rotates itself by driving the two side wheels. Once the required displacement is achieved, it halts one side and propels the robot using the central and outer wheels, thereby correcting the tail's rotation while maintaining the lateral displacement. 

When performing turning locomotion, the LandSalp generates a rotational moment by driving the two side links in opposite directions and then uses the central and head wheels to correct positional errors and shape drift. Notably, during turning, the tail link is pulled by the cable upon reaching the far end, resulting in a significant shape error. In the subsequent gait cycle, feedback control actively adjusts the gait to fix this error.\footnote{Readers interested in further details can refer to the locomotion video provided in the supplementary material.} 

\subsubsection{Shape Regulation Evaluation}

\begin{figure*}[!t]
\centering
\includegraphics[width=\linewidth]{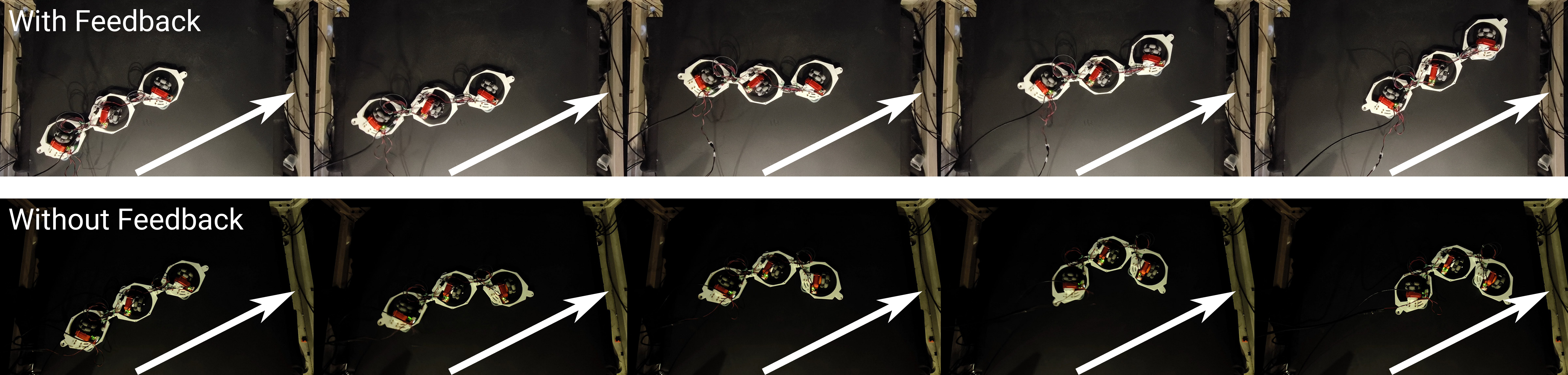}
\caption{Snapshots of LandSalp forward motion with (top) and without (bottom) feedback control. Each row show the robot's positions at approximately the quarter intervals of the entire motion cycle, as well as the configurations corresponding to the quarter intervals of the gait cycle. The white arrows indicate the desired direction of velocity.}
\label{fig:with_without_feedback_snapshot}
\end{figure*}

\begin{figure}[!t]
\centering
\includegraphics[width=\linewidth]{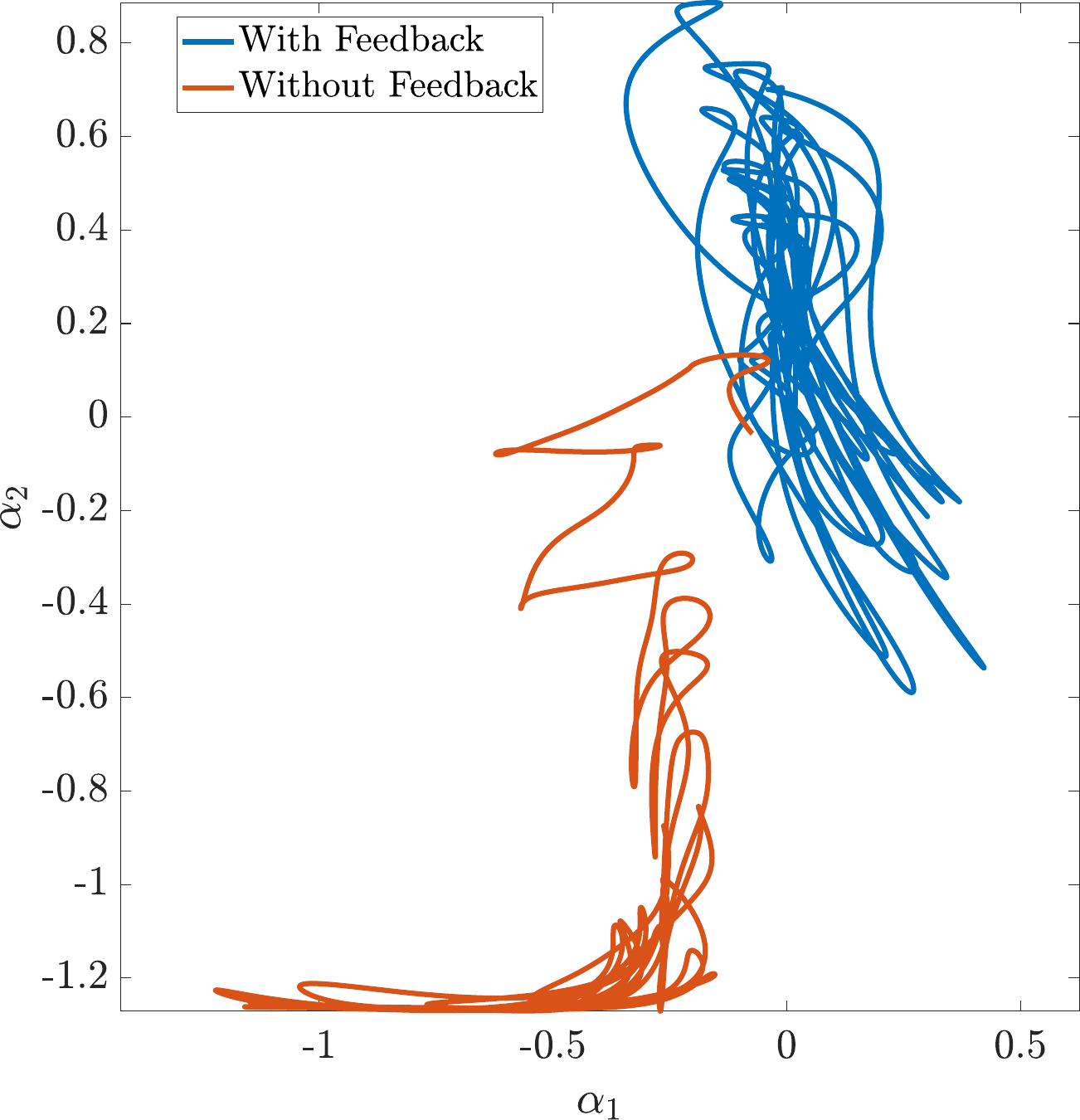}
\caption{Shape trajectory during forward motion with and without feedback control.}
\label{fig:with_without_feedback}
\end{figure}

To emphasize the distinctive challenge of this system---namely, that control is not restricted to the shape axes---we compare axial motion with and without feedback in Fig.~\ref{fig:with_without_feedback_snapshot} and \ref{fig:with_without_feedback}. Without feedback, the robot gradually drifts away from the nominal shape; after several cycles it approaches joint limits and the gait calculated based on averaging about the nominal shape becomes invalid, leading to failure. With feedback, the robot remains near the nominal shape and achieves sustained, accurate locomotion. This comparison demonstrates that the proposed feedback law is essential for preventing shape drift while moving the robot, and thus supports our claims regarding shape controllability.

\subsubsection{Shape Control Evaluation}

\begin{figure}[!t]
\centering
\includegraphics[width=\linewidth]{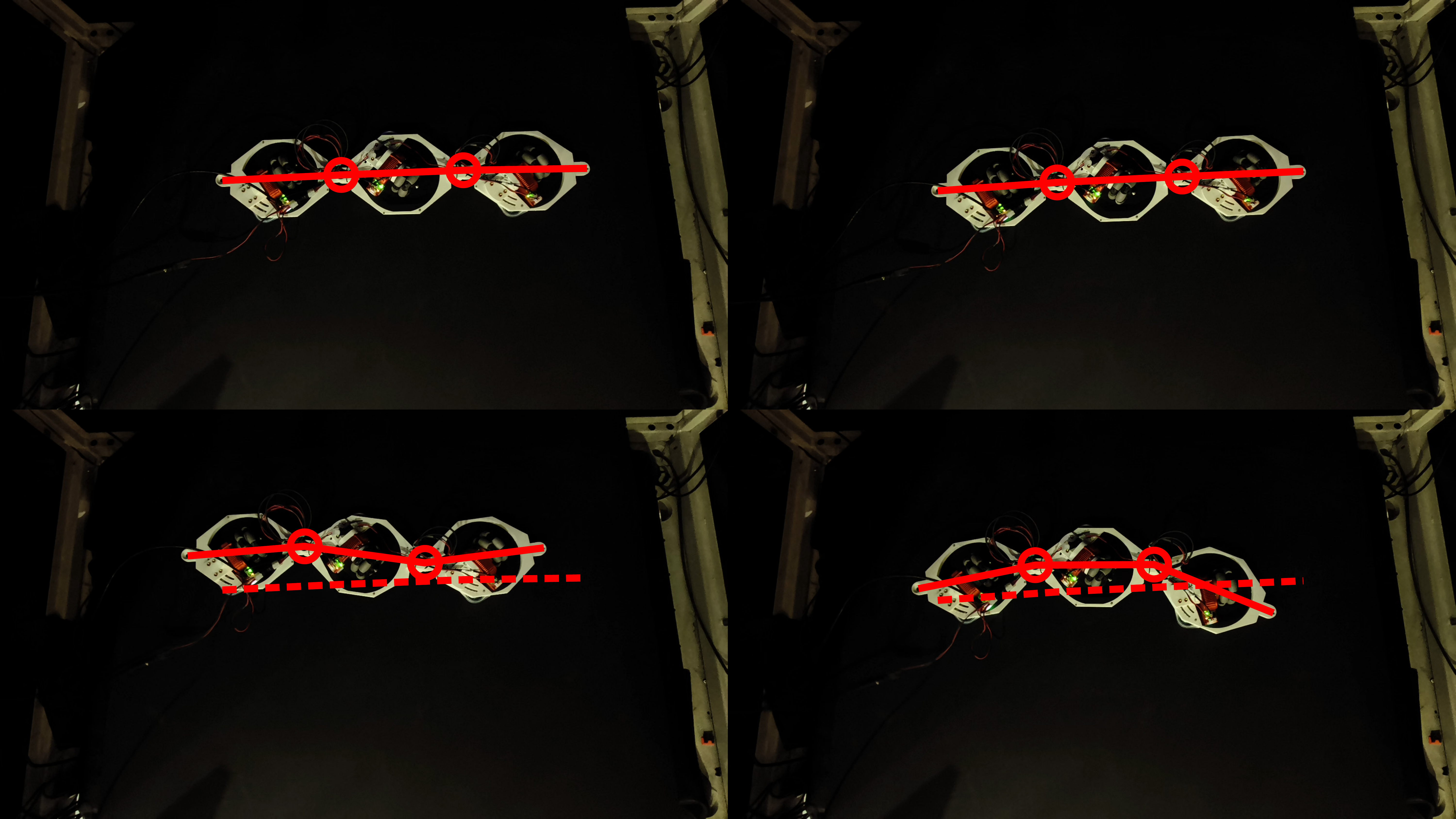}
\caption{Initial and final configurations for the bending experiment, transitioning from the straight configuration (top) to S- and C-shaped configurations (bottom).}
\label{fig:bending_motion}
\end{figure}

We additionally demonstrate a bending maneuver from the straight nominal configuration to C- and S-shaped configurations in Fig.~\ref{fig:bending_motion}. We command a gait at 1~Hz with a target joint-shape velocity of $5/3^\circ$/cycle in the corresponding directions for 12 seconds, which corresponds to a desired bending angle of $\pm 20^\circ$ at each joint. Using the gaits and feedback law designed about the straight nominal shape, the robot reliably achieves the intended bent configurations while regulating position drift. Although these motions move the system away from the nominal shape used for gait design, the robot approaches the desired shapes with acceptable accuracy. This experiment further supports small-time local controllability and indicates a practical basin of attraction about the nominal shape under the proposed framework.

Overall, the optimal gaits successfully achieve accurate omnidirectional locomotion in a reasonable manner within the constraints of actuator capabilities during hardware evaluations. These results demonstrate the effectiveness of the proposed geometric model, data-driven geometric model identification, and geometric motion planning and control methodologies. Furthermore, the locomotion behavior of LandSalp offers valuable insights into the underwater locomotion of salps and provides a path towards the development of more capable salp-inspired underwater robots.

Finally, we emphasize that biological salps and underwater salp-inspired robots are influenced by additional hydrodynamic effects beyond linear viscous drag, including added mass, unsteady vortex shedding, and quadratic drag. In our framework, such effects can be treated as disturbances to be regulated by feedback, or incorporated explicitly in future model extensions. Notably, even on LandSalp, unmodeled effects beyond idealized viscous drag are present (e.g., Coulomb friction, rolling resistance, and cable/harness disturbances). The successful experimental results indicate that the proposed feedback strategy can reject a useful level of unmodeled dynamics and external disturbances, providing evidence that the approach can generalize to underwater robotic platforms operating in regimes where viscous effects are dominant.

\section{Conclusion and Future Work}
\label{sec:conclusion}

This paper developed a comprehensive geometric mechanics framework for analyzing and controlling salp-inspired locomotion, spanning geometric modeling and controllability analysis, data-driven model identification, motion planning, and feedback control validated on the LandSalp robot.

First, we introduced and analyzed a geometric mechanics model of the highly underactuated, drag-dominated, chain-like multi-jet locomotion in which actuation is not restricted to the shape axes. This structure induces coupled, passive shape--position dynamics that differ qualitatively from many canonical geometric locomoting systems. We studied nonlinear controllability and proposed geometric motion planning and feedback control methods tailored to this coupled setting. Using averaging-based analysis and power-based cometrics, we synthesized gaits and feedback controllers that achieve precise omnidirectional locomotion for a desired body velocity and nominal shape while ensuring stability in a force-controlled multi-jet regime.

Second, we extended least-squares and inverse-dynamics system identification techniques to identify the Riemannian metric that encodes a drag-dominated model. Leveraging Lie-group differentiation and a semidefinite programming formulation, we showed that model parameters can be identified from only a few minutes of experimental data. This result highlights both the compactness and expressiveness of the geometric model and provides a general recipe for learning metric structures that capture locomotion dynamics in other bio-inspired robotic platforms.

Finally, we integrated these geometric tools with biological insight into low Reynolds number salp swimming and validated them on LandSalp, a land-based salp analogue with series-elastic actuation. LandSalp serves as a physical realization of a reduced-order salp swimming model, enabling repeatable experiments without the overhead and confounding factors of underwater operation. The agreement between simulation and hardware results supports the effectiveness of our modeling, identification, and control framework and provides a foundation for extending these methods to more complex salp-inspired chains, three-dimensional locomotion, and more challenging environments.

Several directions for future work follow naturally. One immediate direction for future work is adapting our methods to underwater robots propelled by thrusters. This extension would allow us to transfer our insights from the LandSalp to an aquatic environment and tackle the challenges and opportunities of field robotic deployment. Moreover, water-based platforms offer an ideal testbed for evaluating salp-inspired soft jet propulsion units \cite{jones2024underwater}. From a design perspective, part of the modeling and control burden may be shifted to ``mechanical intelligence''~\cite{wang2023mechanical}, for example by introducing joint springs to improve shape stability or anisotropic stiffness/damping to promote obstacle-avoiding behaviors in minimally structured environments.

Beyond the design of the robotic system and jet mechanism, the integration of onboard sensing and state estimation is critical for field deployment. Specifically, inertial measurement units and joint encoders can be used to construct a state estimator based on the proposed locomotion model \cite{rollinson2011state, andrews2025modeling, andrews2025pose}, which is essential for data collection and feedback control. Potential research directions in this domain also include optimizing sensor placement \cite{hinson2014observabilitybased, hinson2015gyroscopic, brace2022sensor} and developing motion planning strategies to enhance information gathering efficacy \cite{hinson2013path, quenzer2014observability, yang2022uav}. 

The onboard state estimation enables the proposed data-driven modeling with only proprioceptive data. This demonstrates the versatility of the proposed approach and suggests future research opportunities that incorporate our methods with onboard sensing to achieve online data-driven model identification \cite{boyacioglu2023nonlinear, sundquist2024nonlinear} and iterative learning control \cite{bittner2018geometrically}, thereby enabling the identification of changes in the swimming environment or the docking and undocking of salp units from the salp robot.

It is important to note that our data-driven modeling approach is not limited to drag-dominated kinematic model. The proposed idea of utilizing the geometric structure to efficiently extract the Riemannian metric that determines the system dynamics from experimental data can be generalized to more general dynamic data-driven models, including inertia with fluid-added mass as well as the stiffness of the salp body \cite{justus2024geometry}. Additional unmodeled effects, potentially arising from vortices and other complex fluid dynamics, can be incorporated by combining the proposed method with the application of Gaussian process \cite{sabelhaus2021gaussian} or Koopman operator theory \cite{bruder2021datadriven}.

As shown in the experimental results, some second-order dynamic effects emerge in the system. This behavior may be more pronounced when we apply the same approach to an underwater robot. To address this, we can apply the methods outlined in Section~\ref{sec:motion_planning} to the dynamic model to find optimal oscillation control based on controllability analysis, similar to \cite{morgansen2007geometric}. To mitigate drift effects in the dynamics, the method discussed in \cite{yang2023geometric} can be used to refine the gait under conditions of non-zero net momentum. 

Finally, another potential avenue for future research involves extending to higher-level path planning that integrates the low-level control proposed in this study. When the command is beyond local velocity and involves waypoints or trajectories, determining the optimal path to track these directives in complex environments with obstacles becomes a challenging problem. This issue can be approached either as a trajectory planning problem that incorporates full dynamics \cite{cabrera2024optimal, choi2024geometric}, or as a hierarchical framework, where we first generate motion primitives or a gait library \cite{choi2022optimal} and subsequently address the path planning problem \cite{hatton2013snakes}. In addition, considering the reconfigurable properties of salps, there are also more potential future works in the decision-making field, such as constructing motion plans for salp-inspired robots with reconfiguration capabilities \cite{casal1999selfreconfiguration}.

\appendices

\section{Group Action and Differentiation}
\label{appendix:group}

In this paper, we take several derivatives with respect to right group actions, producing left-invariant representations of velocity and force. The ``open circle'' oversets represent tangent vectors, such as velocity, and the undersets are for covectors such as force or momentum. Following convention, velocity is represented as the time derivative of the configuration (with the open circle acting in the role of the dot), and momentum and force are given their own symbols (rather than marking them as derivatives of energy terms). 

For example, the full reading of the notation of the body velocity $\fibercirc$ is ``the velocity constructed by taking the derivative of position with respect to right group actions''
\beq
    \fibercirc = \inv{\fiber}\fiberdot,
\eeq
and the full reading of the notation of body force $\coforce$ is ``the force constructed by taking the derivative of dissipated power with respect to right (body) velocities''
\beq
    P = \frac{1}{2}\langle\coforce; \fibercirc\rangle.
\eeq

Similarly, given a time series of group elements, we can apply Lie-group time differentiation to obtain a discrete approximation of the body velocity in the corresponding Lie algebra, which better aligns with the underlying geometric structure:
\beq
    \fibercirc(t) = \frac{1}{\Delta t}\log\left(\inv{\fiber(t)}\fiber(t+\Delta t)\right).
\eeq
Unlike finite differencing in Euclidean space, which takes the underlying trajectory as linearly interpolating between evaluation points, Lie-group differentiation uses an arc interpolation which naturally captures the intrinsic coupling between translation and rotation for systems whose dynamics are expressed in their body frames.

\section{Use Averaging Theory for Shape-Controlled System}
\label{appendix:shape}

In this paper, we consider systems where forces serve as the controlled input. It is noteworthy that the same analytical framework can be extended to systems with shape control. The application of averaging theory in the locomotion analysis of these systems offers significant insights into geometric motion planning.

Suppose that we can directly control shape velocity to achieve locomotion, i.e. $\fibercirc = \cvf(\base)\basedot$, which is a common setting in many previous studies on geometric motion planning \cite{hatton2011geometric, hatton2015nonconservativity, ramasamy2019geometry, hatton2022geometry}. The averaging process outlined in~\eqref{eq:liebracket} is still applicable and becomes
\begin{eqalign}
    \fibercirc^{\mathrm{avg}} T &= \log{\fiber(T)} \\
    &= \log\left(\Prodi^{T}_{0}\exp\left(\cvf\basedot(t) dt\right)\right)\\
    &= \int^{T}_{0}\cvf\basedot(t) dt \nonumber \\
    &\quad+ \int^{T}_{0}\frac{1}{2}\left[\cvf\basedot(t), \left(\int^{T}_{t}\cvf\basedot(\tau)d\tau\right)\right] dt + \cdots \\
    &\approx \int^{T}_{0}\cvf\basedot(t) dt \nonumber \\
    &\quad+ \int^{T}_{0}\frac{1}{2}\left[\cvf\basedot(t), \left(\int^{T}_{t}\cvf\basedot(\tau)d\tau\right)\right] dt \\
    &= \oiint \left(\extd\cvf + \sum_{i, j>i}[\cvf_i, \cvf_{j}]\right),
\end{eqalign}
where the generalized Stokes' theorem is applied to the line integral, and the one-half factor is canceled by the antisymmetric property of the Lie brackets, resulting in integrals over 2-surfaces in the shape space. It is important to highlight that there is no necessity to assume a constant control vector field at the nominal shape in this case, because we are directly working within the shape space and integrating the control vector field therein. This result aligns with the total Lie bracket described in the previous geometric mechanics literature \cite{hatton2015nonconservativity}, and can be extended for higher-order approximations as discussed in~\cite{bass2022characterizing}.

\section{Gait Optimization Considering Average Shape and Joint Limits}
\label{appendix:average}

As demonstrated by the shape trajectories observed in the simulation evaluations, for certain nominal shapes far from the zero joint angles, the shape trajectory within a single gait cycle can exhibit substantial deviations from the nominal shape and approach the joint limits. Given that the linearization is performed around the nominal shape, such large deviations can invalidate the feedback control. In addition, considering the joint limitations of the robot, large joint motions can potentially result in collisions between links.

To address this issue, a possible solution involves incorporating additional decision variables and constraints into the motion planning optimization to ensure that the nominal shape is the average of the shape trajectories throughout the gait cycle. Specifically, we can introduce a new variable, $\base_{0}$, representing the initial shape of the gait cycle, rather than assuming that the gait starts and ends with the nominal shape. Additionally, it is necessary to impose a constraint such that the mean of the shape trajectories within the gait cycle is the nominal shape. The new optimization problem can be formulated as follows:
\begin{eqalign}
    \min_{\Ctrl, \base_{0}} \quad& \transpose{\Ctrl}\costmetric\Ctrl \\
    s.t. \quad& \frac{1}{T}\log\fiber(T) - \bundledot^{\mathrm{desired}} = 0 \\
    & \frac{1}{T}\int^{T}_{0}\base(t)dt = \base^{\mathrm{nominal}} \\
    & \base(t) \in \mathcal{R} \quad \forall t,
\end{eqalign}
where $\mathcal{R}$ is the joint limit.

The new optimization formula better guarantees that the optimal gait converges to the nominal shape on average, thereby ensuring linearization performance while minimizing the risk of the range of motion exceeding joint limits. However, this modification increases the complexity of the optimization. In particular, the solution for the constrained optimization can differ significantly from that obtained using an approximate optimization based on the averaged system, complicating any preliminary rough search for the optimal gait. Given the challenge of analytically obtaining the gradient in the full optimization and the inefficiency of finite-difference methods, this optimization may not be feasible. Additionally, this formula imposes higher demands on the feedback control, because the shape trajectory is constrained to average to the nominal shape rather than passing directly through it. One possible feedback strategy is to integrate the shape online over the gait cycle and use the difference between the integrated average and the nominal shape as the error signal in the feedback loop. In practice, when operating near the zero joint angles, the shape trajectory span is relatively small, so it is not necessary to use this formula. In future work, this new formula can be used for more demanding tasks.

Another approach is to use the optimal gait from the averaged system directly, rather than refining the gait on the full system to force the shape trajectory to pass through the nominal shape. This approach requires modifying the feedback controller to account for the fact that the nominal shape corresponds to the cycle-average shape, rather than the shape at the initial phase. This modification can be done by replacing~\eqref{eq:feedback} with
\begin{eqalign}
    \Ctrl(t) &= \Ctrl_0 - \feedback\Delta\bundle(t_\trdidx) \\
    \Delta\bundle(t_\trdidx) &= \frac{1}{T}\int^{t_{\trdidx}}_{t_\trdidx-1} \left(\bundle^{\mathrm{desired}}(t) - \bundle(t)\right) dt \nonumber \\
    &\text{for } t\in[t_\trdidx, t_{\trdidx+1}), \text{where } t_\trdidx=\trdidx T, \trdidx\in\mathbb{Z}_{\ge 0},
\end{eqalign}
so that feedback is based on the integrated configuration error over the previous cycle. This approach enables regulation of average shape error and can help maintain stability about the nominal operating point. Because the optimal gait from the averaged system is used directly, this method naturally targets trajectories whose cycle-average shape matches the nominal shape. In hardware, however, it requires sufficiently accurate timing and numerical integration over each gait cycle.

\section{Averaging Accuracy and Coordinate Optimization}
\label{appendix:coordinate}

As previously noted, the accuracy of the average approximation in~\eqref{eq:liebracket} depends on the proportion of the truncated higher-order terms in the BCH series. To enable the use of gaits with low frequency and large control input, it is essential to maintain approximation accuracy. Because all higher-order terms involve higher-order Lie brackets of the control vector field $\cvf$, to improve accuracy, we can identify the optimal coordinates with transformation $\coordtrans(\base)$ relative to the original coordinates. This transformation aims to minimize the magnitude of the control vector field over a region of interest $\coordoptroi$ in the shape space, thereby minimizing the proportion of the truncated higher-order terms
\begin{eqalign}
    \cvf_{\mathrm{new}}&=\begin{bmatrix}\inv{\Adj_{\coordtrans}}&0\\0&\matrixid\end{bmatrix}\cvf + \begin{bmatrix}\coordtranscirc \\ 0\end{bmatrix} \label{eq:cvfcoord} \\
    \coordtrans^{*} &= \argmin_{\coordtrans} \idotsint_{\base \in \coordoptroi}\left\|\cvf_{\mathrm{new}}(\base)\right\|^{2}_{F} d\coordoptroi.
\end{eqalign}
The above minimization problem is a partial differential equation. Previous works have introduced two different methods for determining the optimal coordinates: one involves discretizing the shape space into a grid \cite{hatton2011geometric}, and the other entails sampling along the gait within the shape space \cite{yang2024geometric}. Solutions to the minimization tend to place the system body frame near its center of mass and the mean orientation of the links; in many cases sufficient accuracy can be achieved by using such an averaged frame directly instead of performing the optimization.

It is worth noting that changing the coordinates does not affect the magnitude of the control vector field in the direction of the shape, as shown in~\eqref{eq:cvfcoord}, which demonstrates a key distinction between force control systems and shape control systems. Therefore, the stability of a control scheme based on averaging theory will depend on the variation of the control vector field around the nominal shape. One possible direction is to consider the variations in the control vector field during motion planning, rather than assuming the shape remains constant at its nominal value.

\bibliographystyle{IEEEtran}
\bibliography{ref}

\end{document}